\documentclass[final]{elsarticle}
\usepackage{times,amsmath,amssymb,subfigure,lineno,hyperref,url,multirow}
\usepackage[flushleft]{threeparttable}
\usepackage[lined,ruled,commentsnumbered]{algorithm2e}
\allowdisplaybreaks

\begin{document}

\begin{frontmatter}
\title{Active Learning for Regression Using Greedy Sampling}

\author[HUST]{Dongrui Wu\corref{CA}}\ead{drwu@hust.edu.cn}

\author[UTS]{Chin-Teng Lin}\ead{Chin-Teng.Lin@uts.edu.au}

\author[HUST]{Jian Huang\corref{CA}}\cortext[CA]{Corresponding author}\ead{huang\_jan@mail.hust.edu.cn}

\address[HUST]{Key Laboratory of the Ministry of Education for Image Processing and Intelligent Control,\\ School of Automation, Huazhong University of Science and Technology, Wuhan 430074, China}

\address[UTS]{Faculty of Engineering and Information Technology, University of Technology, Sydney, Australia}

\begin{abstract}
Regression problems are pervasive in real-world applications. Generally a substantial amount of labeled samples are needed to build a regression model with good generalization ability. However, many times it is relatively easy to collect a large number of unlabeled samples, but time-consuming or expensive to label them. Active learning for regression (ALR) is a methodology to reduce the number of labeled samples, by selecting the most beneficial ones to label, instead of random selection. This paper proposes two new ALR approaches based on greedy sampling (GS). The first approach (GSy) selects new samples to increase the diversity in the output space, and the second (iGS) selects new samples to increase the diversity in both input and output spaces. Extensive experiments on 12 UCI and CMU StatLib datasets from various domains, and on 15 subjects on EEG-based driver drowsiness estimation, verified their effectiveness and robustness.
\end{abstract}

\end{frontmatter}


\section{Introduction}

Regression, which estimates the value of a dependent variable (output) from one or more independent variables (predictors, features, inputs), is a common problem in machine learning. To build an accurate regression model, one needs to have some labeled training samples, i.e., samples whose dependent and independent variable values are both known. Generally the more the labeled training samples are, the better the regression performance is. However, in real-world many times it is relatively easy to obtain the values of the inputs, but time-consuming or expensive to label them. For example, in speech emotion estimation \cite{drwuICME2010,drwuInterSpeech2010} in the 3-dimensional space of valance, arousal and dominance \cite{Mehrabian1980}, it is easy to record a large number of voice pieces, but time-consuming to evaluate their emotions \cite{Grimm2008,Bradley2007}. Another example is driver drowsiness estimation from physiological signals such as the electroencephalogram (EEG) \cite{drwuTFS2017,drwuEBMAL2016,drwuSMLR2016}. It is relatively easy to collect a large number of EEG trials, but not easy to obtain their groundtruth drowsiness.

There are at least three different directions for performing regression based on a small number of labeled training samples:
\begin{enumerate}
\item \emph{Regularization} \cite{Zou2005,Hastie2009}, which introduces additional information to improve the generalization performance. For example, ridge regression (RR) \cite{Hoerl1970} and LASSO \cite{Tibshirani1996} penalize large regression coefficients.
\item \emph{Transfer learning} \cite{Pan2010}, which uses data or information from related domains/tasks to improve the regression performance. For example, labeled EEG data from other subjects could be used to improve the drowsiness estimation performance for a new subject \cite{drwuTFS2017,drwuaBCI2015}.
\item \emph{Active learning} \cite{Settles2009}, which selects the most beneficial unlabeled samples to label, instead of random selection. For example,  batch-mode active learning has been employed for EEG-based driver drowsiness estimation \cite{drwuEBMAL2016}.
\end{enumerate}

This paper focuses on the third direction, i.e., active learning for regression (ALR). Particularly, we consider sequential pool-based ALR \cite{Sugiyama2009}, in which a pool of unlabeled samples is given, and the goal is to sequentially choose some to label, so that a regression model trained from them can give the most accurate estimates for the remaining unlabeled samples.

Compared with the large literature on active learning for classification, there are only a few approaches for sequential pool-based ALR \cite{Burbidge2007,Cai2013,Yu2010}. This paper proposes two new ALR approaches, inspired by the greedy sampling (GS) approach in \cite{Yu2010}. Extensive experiments on 12 UCI and CMU StatLib datasets from various domains, and on 15 subjects on EEG-based driver drowsiness estimation, verified their effectiveness and robustness.

The remainder of this paper is organized as follows: Section~\ref{sect:GSALR} introduces the original GS ALR approach, and proposes two new ALR algorithms. Section~\ref{sect:UCI} describes the 12 UCI and CMU StatLib datasets for evaluating the effectiveness of different ALR approaches, and the corresponding experimental results. Section~\ref{sect:Driving} describes the offline EEG-based driver drowsiness estimation experiment for evaluating the effectiveness of different ALR approaches, and the corresponding experimental results. Finally, Section~\ref{sect:conclusions} draws conclusions and points out some future research directions.

\section{Greedy Sampling ALR Approaches} \label{sect:GSALR}

In this section we introduce an existing GS ALR approach in the literature, and propose two new ALR approaches.

\subsection{Greedy Sampling on the Inputs (GSx)} \label{sect:GS}

Yu and Kim \cite{Yu2010} proposed four passive sampling approaches for regression. Different from most ALR approaches, which generally require updating the regression model in each iteration and computing the predictions for the unlabeled samples, passive sampling selects the sample based entirely on its location in the feature space. Thus, it is independent of the regression model, and has low computational cost.

Among the several passive sampling approaches in \cite{Yu2010}, GS achieved the best overall performance. However, the original GS approach did not explain how the first sample was selected. This subsection introduces GSx, which is essentially the same as GS, except that it also includes a strategy to select the first sample for labeling.

Assume the pool consists of $N$ samples $\{\mathbf{x}_n\}_{n=1}^N$, initially none of which is labeled. Our goal is to select $K$ of them to label, and then construct an accurate regression model from them to estimate the outputs for the remaining $N-K$ samples. GSx selects the first sample as the one closest to the centroid of all $N$ samples (i.e., the one with the shortest distance to the remaining $N-1$ samples), and the remaining $K-1$ samples incrementally. The idea is to make the first selection most representative.

Without loss of generality, assume the first $k$ samples have already been selected. For each of the remaining $N-k$ unlabeled samples $\{\mathbf{x}_n\}_{n=k+1}^N$, GSx computes first its distance to each of the $k$ labeled samples:
\begin{align}
d_{nm}^{\mathbf{x}}=||\mathbf{x}_n-\mathbf{x}_m||,\quad m=1,...,k; n=k+1,...,N \label{eq:dnmx}
\end{align}
then $d_n^{\mathbf{x}}$, the shortest distance from $\mathbf{x}_n$ to all $k$ labeled samples:
\begin{align}
d_n^{\mathbf{x}}=\min_m d_{nm}^{\mathbf{x}},\quad n=k+1,...,N \label{eq:dnx}
\end{align}
and finally selects the sample with the maximum $d_n^{\mathbf{x}}$ to label.

In summary, GSx selects the first sample as the one closest to the centroid of the pool, and in each subsequent iteration a new sample located furthest away from all previously selected samples in the input space to achieve the diversity among the selected samples. Its pseudo-code is given in Algorithm~\ref{alg:GSx}.

\begin{algorithm}[!h]
\KwIn{$N$ unlabeled samples, $\{\mathbf{x}_n\}_{n=1}^N$\;
\hspace*{10mm} $K$, the maximum number of labels to query.}
\KwOut{The regression model $f(\mathbf{x})$.}
\tcp{Initialize the first selection}
Set $Z=\{\mathbf{x}_n\}_{n=1}^N$, and $S=\emptyset$\;
Identify $\mathbf{x}'$, the sample closest to the centroid of $Z$\;
Move $\mathbf{x}'$ from $Z$ to $S$\;
Re-index the sample in $S$ as $\mathbf{x}_1$, and the samples in $Z$ as $\{\mathbf{x}_n\}_{n=2}^N$\;
\tcp{Select $K-1$ more samples incrementally}
\For{$k=1,...,K-1$}{
\For{$n=k+1,...,N$}{
Compute $d_n^{\mathbf{x}}$ in (\ref{eq:dnx})\;}
Identify the $\mathbf{x}'$ that has the largest $d_n^{\mathbf{x}}$\;
Move $\mathbf{x}'$ from $Z$ to $S$\;
Re-index the samples in $S$ as $\{\mathbf{x}_m\}_{m=1}^{k+1}$, and the samples in $Z$ as $\{\mathbf{x}_n\}_{n=k+2}^N$\;}
Query to label all $K$ samples in $S$\;
Construct the regression model $f(\mathbf{x})$ from $S$.
\caption{The GSx ALR approach, slightly modified from GS in \cite{Yu2010} on the initialization.} \label{alg:GSx}
\end{algorithm}

\subsection{Greedy Sampling on the Output (GSy)} \label{sect:GSy}

GSx achieves diversity in the input space. Our proposed GSy aims to achieve diversity in the output space.

Like GSx, in GSy initially the pool consists of $N$ unlabeled samples and zero labeled sample. To evaluate the diversity in the output space, we need to know the outputs (labels) of all samples, either true or estimated. In other words, GSy cannot be applied before $K_0$ labeled samples are obtained, where $K_0$ is the minimum number of labeled samples required to build a regression model. In this paper we set $K_0$ as the number of features in the input space, and use GSx to select the first $K_0$ samples to label.

Assume the first $k$ ($k\ge K_0$) samples have already been labeled with outputs $\{y_m\}_{m=1}^k$, and a regression model $f(\mathbf{x})$ has been constructed. For each of the remaining $N-k$ unlabeled sample $\{\mathbf{x}_n\}_{n=k+1}^N$, GSy computes first its distance to each of the $k$ outputs:
\begin{align}
d_{nm}^y=||f(\mathbf{x}_n)-y_m||,\quad m=1,...,k; n=k+1,...,N \label{eq:dnmy}
\end{align}
and $d_n^y$, the shortest distance from $f(\mathbf{x}_n)$ to $\{y_m\}_{m=1}^k$:
\begin{align}
d_n^y=\min_m d_{nm}^y,\quad n=k+1,...,N \label{eq:dny}
\end{align}
and then selects the sample with the maximum $d_n^y$ to label.

In summary, GSy selects the first a few samples using GSx to build an initial regression model, and then in each subsequent iteration a new sample located furthest away from all previously selected samples in the output space to achieve diversity among the selected samples. Its pseudo-code is given in Algorithm~\ref{alg:GSy}. Note that GSy is no longer a passive sampling approach, because it needs to update $f(\mathbf{x})$ in each iteration.

\begin{algorithm}[!h] 
\KwIn{$N$ unlabeled samples, $\{\mathbf{x}_n\}_{n=1}^N$\;
\hspace*{10mm} $K$, the maximum number of labels to query.}
\KwOut{The regression model $f(\mathbf{x})$.}
\tcp{Initialize the first selection}
Set $Z=\{\mathbf{x}_n\}_{n=1}^N$, and $S=\emptyset$\;
Identify $\mathbf{x}'$, the sample closest to the centroid of $Z$\;
Move $\mathbf{x}'$ from $Z$ to $S$\;
Re-index the sample in $S$ as $\mathbf{x}_1$, and the samples in $Z$ as $\{\mathbf{x}_n\}_{n=2}^N$\;
\tcp{Select $K_0-1$ more samples incrementally using GSx}
Identify $K_0$, the minimum number of labeled samples required to construct $f(\mathbf{x})$\;
\For{$k=1,...,K_0-1$}{
\For{$n=k,...,N$}{
Compute $d_n^{\mathbf{x}}$ in (\ref{eq:dnx})\;}
Identify the $\mathbf{x}'$ that has the largest $d_n^{\mathbf{x}}$\;
Move $\mathbf{x}'$ from $Z$ to $S$\;
Re-index the samples in $S$ as $\{\mathbf{x}_m\}_{m=1}^{k+1}$, and the samples in $Z$ as $\{\mathbf{x}_n\}_{n=k+2}^N$\;}
Query to label the $K_0$ samples in $S$\;
Construct the regression model $f(\mathbf{x})$ from $S$\;
\tcp{Select $K-K_0$ more samples incrementally}
\For{$k=K_0,...,K-1$}{
\For{$n=k,...,N$}{
Compute $d_n^y$ in (\ref{eq:dny})\;}
Identify the $\mathbf{x}'$ that has the largest $d_n^y$\;
Move $\mathbf{x}'$ from $Z$ to $S$\;
Query to label $\mathbf{x}'$ in $S$\;
Re-index the samples in $S$ as $\{\mathbf{x}_m\}_{m=1}^{k+1}$, and the samples in $Z$ as $\{\mathbf{x}_n\}_{n=k+2}^N$\;
Update the regression model $f(\mathbf{x})$ using $S$.}
\caption{The GSy ALR approach.} \label{alg:GSy}
\end{algorithm}
\renewcommand{\baselinestretch}{1.5}

\begin{figure}[!h]\centering
\includegraphics[width=.8\linewidth,clip]{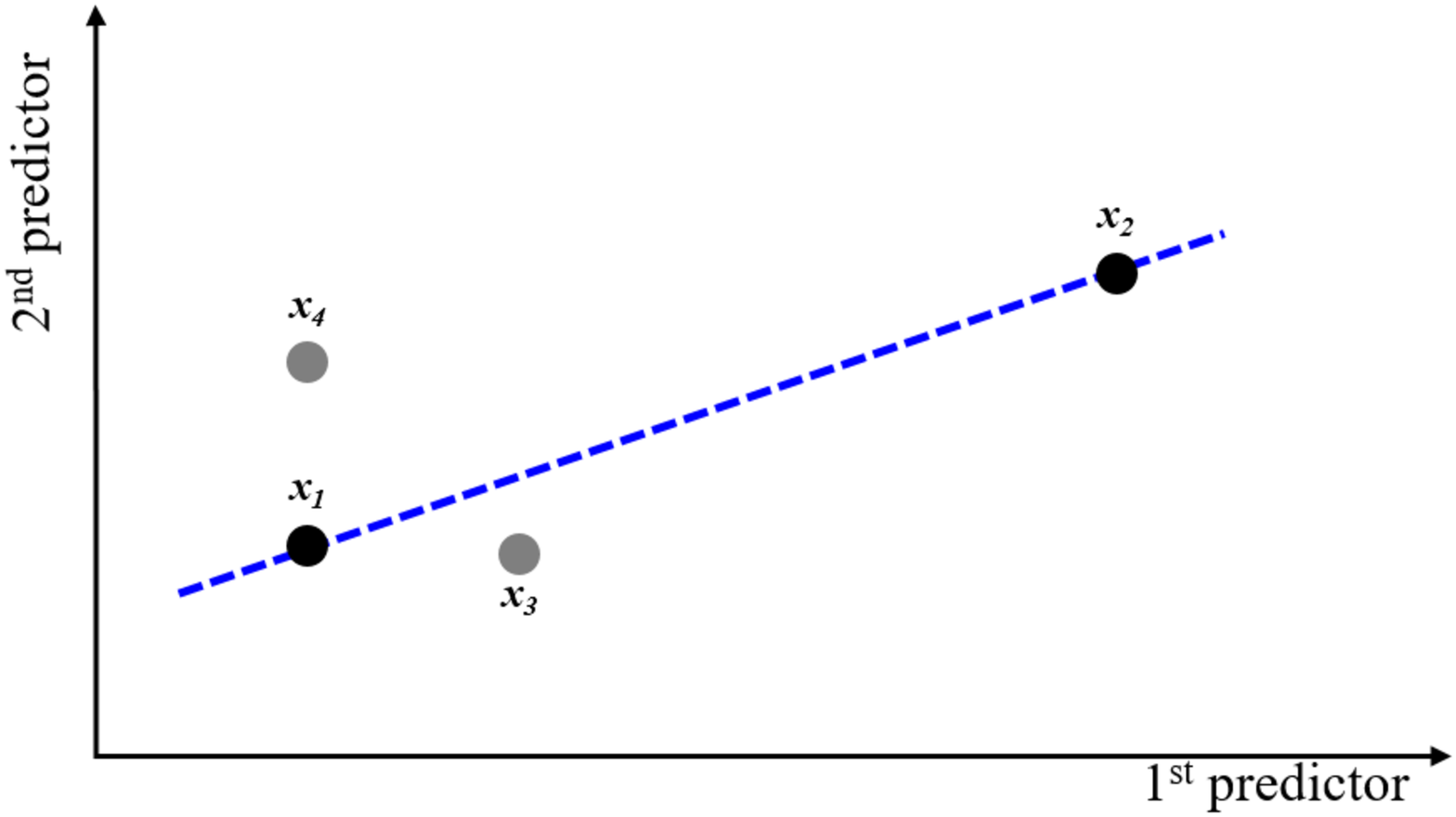}
\caption{Illustration of GSy. GSy will select $\mathbf{x}_3$ to label if the 1st predictor is more sensitive (important) than the 2nd, and $\mathbf{x}_4$ otherwise.} \label{fig:GSy}
\end{figure}

The rationale for GSy can be illustrated by the following simple example shown in Figure~\ref{fig:GSy}. Assume the input space has only two dimensions, and we have only four samples:
\begin{align}
\mathbf{x}_1&=(x_{11},x_{12})^T\\
\mathbf{x}_2&=(x_{21},x_{22})^T\\
\mathbf{x}_3&=(x_{31},x_{32})^T=(x_{11}+\delta,x_{12})^T\\
\mathbf{x}_4&=(x_{41},x_{42})^T=(x_{11},x_{12}+\delta)^T
\end{align}
where the first two have labels $y_1$ and $y_2$, respectively, the last two are unlabeled, and $\delta$ is a small number. A regression function $f(\mathbf{x})$ is built from $(\mathbf{x}_1,y_1)$ and $(\mathbf{x}_2,y_2)$. We want to select $\mathbf{x}_3$ or $\mathbf{x}_4$ to label so that the estimation error of $f(\mathbf{x})$ on the four samples can be maximally reduced.

For simplicity, assume both $f(\mathbf{x}_3)$ and $f(\mathbf{x}_4)$ are closer to $y_1$ than to $y_2$, so we only need to consider their distances to $y_1$ in GSy. The sensitivity of $f(\mathbf{x})$ to the first predictor, evaluated around $\mathbf{x}_1$, can be approximated as
\begin{align}
s_1=\frac{|f(\mathbf{x}_3)-y_1|}{|x_{31}-x_{11}|}=\frac{|f(\mathbf{x}_3)-y_1|}{|\delta|}
\end{align}
Similarly, the sensitivity of $f(\mathbf{x})$ to the second predictor, evaluated also around $\mathbf{x}_1$, can be approximated as
\begin{align}
s_2=\frac{|f(\mathbf{x}_4)-y_1|}{|x_{42}-x_{12}|}=\frac{|f(\mathbf{x}_4)-y_1|}{|\delta|}
\end{align}
When $|s_1|>|s_2|$, which means $f(\mathbf{x})$ is more sensitive to the first predictor than to the second, we should select a sample that can help refine the first regression coefficient for labeling. Generally the more diverse the values of the first predictor are, the more accurate its regression coefficient can be determined. So, in this case we should select $\mathbf{x}_3$ for labeling. Note that $|s_1|>|s_2|$ implies $|f(\mathbf{x}_3)-y_1|>|f(\mathbf{x}_4)-y_1|$. According to the procedure of GSy, indeed $\mathbf{x}_3$ will be selected. Similarly, when $|s_1|<|s_2|$, GSy will correctly select $\mathbf{x}_4$ for labeling.

\subsection{Improved Greedy Sampling (iGS) on both Inputs and Output} \label{sect:iGS}

GSx considers only the diversity in the input (feature) space, by computing the minimum distance between an unlabeled sample and all existing labeled samples, using all features. However, maybe not all features are useful; even if all features are useful, they may have different importance. GSx does not take feature selection/weighting into consideration.

GSy considers only the diversity in the output (label) space, by computing the minimum distance between the estimated output for a sample and all existing outputs. Our example in the previous subsection shows that GSy tries to select a new sample that can significantly increase the diversity of the most sensitive predictor. Thus, it implicitly considers feature selection/weighting. However, the predictor sensitivities are evaluated using $f(\mathbf{x})$ constructed from a very small number of labeled samples, so they may not be very accurate. In other words, feature selection/weighting in GSy may not be reliable.

In this subsection we propose an improved greedy sampling (iGS) approach, which combines GSx and GSy, to ensure that we take feature selection/weighting into consideration, but can also avoid catastrophic failure if feature selection/weighting is misleading.

Like GSx and GSy, initially the pool consists of $N$ unlabeled samples and zero labeled sample. In iGS we again set $K_0$ to be the number of features in the input space, and use GSx to select the first $K_0$ samples to label. Assume the first $k$ samples have already been labeled with labels $\{y_n\}_{n=1}^k$. For each of the remaining $N-k$ unlabeled sample $\{\mathbf{x}_n\}_{n=k+1}^N$, iGS computes first $d_{nm}^{\mathbf{x}}$ in (\ref{eq:dnmx}) and $d_{nm}^y$ in (\ref{eq:dnmy}), and $d_n^{\mathbf{x}y}$:
\begin{align}
d_n^{\mathbf{x}y}=\min_m d_{nm}^{\mathbf{x}}d_{nm}^y,\quad n=k+1,...,N \label{eq:dnxy}
\end{align}
and then selects the sample with the maximum $d_n^{\mathbf{x}y}$ to label.

In summary, iGS selects the first a few samples using GSx to build an initial regression model, and then in each subsequent iteration a new sample located furthest away from all previously selected samples in both input and output spaces to achieve balanced diversity among the selected samples. Its pseudo-code is given in Algorithm~\ref{alg:iGS}. Note that iGS is no longer a passive sampling approach, because it needs to update $f(\mathbf{x})$ in each iteration.

\begin{algorithm}[!h] 
\KwIn{$N$ unlabeled samples, $\{\mathbf{x}_n\}_{n=1}^N$\;
\hspace*{10mm} $K$, the maximum number of labels to query.}
\KwOut{The regression model $f(\mathbf{x})$.}
\tcp{Initialize the first selection}
Set $Z=\{\mathbf{x}_n\}_{n=1}^N$, and $S=\emptyset$\;
Identify $\mathbf{x}'$, the sample closest to the centroid of $Z$\;
Move $\mathbf{x}'$ from $Z$ to $S$\;
Re-index the sample in $S$ as $\mathbf{x}_1$, and the samples in $Z$ as $\{\mathbf{x}_n\}_{n=2}^N$\;
\tcp{Select $K_0-1$ more samples incrementally using GSx}
Identify $K_0$, the minimum number of labeled samples required to construct $f(\mathbf{x})$\;
\For{$k=1,...,K_0-1$}{
\For{$n=k,...,N$}{
Identify the $\mathbf{x}'$ that has the largest $d_n^{\mathbf{x}}$\;}
Move $\mathbf{x}'$ from $Z$ to $S$\;
Re-index the samples in $S$ as $\{\mathbf{x}_m\}_{m=1}^{k+1}$, and the samples in $Z$ as $\{\mathbf{x}_n\}_{n=k+2}^N$\;}
Query to label the $K_0$ samples in $S$\;
Construct the regression model $f(\mathbf{x})$ from $S$\;
\tcp{Select $K-K_0$ more samples incrementally}
\For{$k=K_0,...,K-1$}{
\For{$n=k,...,N$}{
Compute $d_n^{\mathbf{x}y}$ in (\ref{eq:dnxy})\;}
Identify the $\mathbf{x}'$ that has the largest $d_n^{\mathbf{x}y}$\;
Move $\mathbf{x}'$ from $Z$ to $S$\;
Query to label $\mathbf{x}'$ in $S$\;
Re-index the samples in $S$ as $\{\mathbf{x}_m\}_{m=1}^{k+1}$, and the samples in $Z$ as $\{\mathbf{x}_n\}_{n=k+2}^N$\;
Update the regression model $f(\mathbf{x})$ using $S$.}
\caption{The iGS ALR approach.} \label{alg:iGS}
\end{algorithm}
\renewcommand{\baselinestretch}{1.5}

\section{Experiments on UCI and CMU StatLib Datasets} \label{sect:UCI}

Extensive experiments on 12 UCI and CMU StatLib datasets are performed in this section to demonstrate the performances of GSy and iGS.

\subsection{Datasets}

We used 12 datasets from the UCI Machine Learning Repository\footnote{\url{http://archive.ics.uci.edu/ml/index.php}} and the CMU StatLib Datasets Archive\footnote{\url{http://lib.stat.cmu.edu/datasets/}} that have been used in previous ALR experiments \cite{Cai2013,Cai2017,Yu2010,drwuSAL2018}. Their summary is given in Table~\ref{tab:datasets}. We used one-hot coding to convert categorical features into numerical features. Then, we normalized each dimension of the feature space to have mean zero and standard deviation one.

\begin{table}[!h]
\caption{Summary of the 12 UCI and CMU StatLib datasets.} \label{tab:datasets}
\centering \setlength{\tabcolsep}{1mm}
\begin{threeparttable}
\begin{tabular}{l|cccccc}   \hline
 & &  No. of         &  No. of    &   No. of   &   No. of  & No. of           \\
Dataset & Source& samples        & raw   &   numerical  &   categorical  & total           \\
& &  &  features  &   features &   features &  features       \\ \hline
Concrete-CS\tnote{1}  &  UCI & 103  &   7 &7 &0 & 7 \\
Concrete-Flow\tnote{1}  &  UCI & 103  &   7 &7 &0 & 7 \\
Concrete-Slump\tnote{1}\hspace*{1mm}  &  UCI &  103  &   7 &7 &0 & 7 \\
Yacht\tnote{2}  &  UCI &   308  &   6&6 & 0& 6\\
autoMPG\tnote{3}  & UCI &   392 & 7  &    6     & 1 & 9 \\
NO2\tnote{4}  &  StatLib &   500  &     7 & 7 & 0   &  7  \\
PM10\tnote{4}  &  StatLib &   500  &     7 & 7 & 0   &  7  \\
Housing\tnote{5} &  UCI & 506 &  13 & 13 & 0    &  13 \\
CPS\tnote{6}  &   StatLib & 534  &   11 & 8 &3     &  19 \\
Concrete\tnote{7}& UCI & 1030 & 8 & 8 & 0 & 8  \\
Wine-red\tnote{8} & UCI & 1599  &  11 & 11 & 0    &   11 \\
Wine-white\tnote{8} & UCI & 4898 &  11 & 11 & 0    &   11 \\  \hline
\end{tabular}
  \begin{tablenotes} \small
\item[1] \url{https://archive.ics.uci.edu/ml/datasets/Concrete+Slump+Test}
\item[2] \url{https://archive.ics.uci.edu/ml/datasets/Yacht+Hydrodynamics}
\item[3] \url{https://archive.ics.uci.edu/ml/datasets/auto+mpg}
\item[4] \url{http://lib.stat.cmu.edu/datasets/}
\item[5] \url{https://archive.ics.uci.edu/ml/machine-learning-databases/housing/}
\item[6] \url{http://lib.stat.cmu.edu/datasets/CPS_85_Wages}
\item[7] \url{https://archive.ics.uci.edu/ml/datasets/Concrete+Compressive+Strength}
\item[8] \url{https://archive.ics.uci.edu/ml/datasets/Wine+Quality}
  \end{tablenotes}
\end{threeparttable}
\end{table}
\renewcommand{\baselinestretch}{1.5}

\subsection{Algorithms} \label{sect:algs}

We compared the performances of six different sample selection algorithms:
\begin{enumerate}
\item Baseline (\texttt{BL}), which randomly selects all $K$ samples.
\item Query-by-Committee (\texttt{QBC}) \cite{RayChaudhuri1995}. It first bootstraps the $k$ labeled samples into $P$ copies, each containing $k$ samples but with duplicates, and builds a regression model from each copy, i.e., the committee consists of $P$ regression models. Then, for each of the $N-k$ unlabeled samples, it computes the variance of the $P$ individual predictions, and selects the one with the maximum variance to label.
\item Expected model change maximization (\texttt{EMCM}) \cite{Cai2013}. It first uses all $k$ labeled samples to build a linear regression model. Then, it also uses bootstrap to construct $P$ linear regression models. For each of the $N-k$ unlabeled samples, it computes the expected model change when that sample is labeled and added to the training dataset. EMCM selects the sample with the maximum expected model change to label.
\item \texttt{GSx}, which has been introduced in Section~\ref{sect:GS}.
\item \texttt{GSy}, which has been introduced in Section~\ref{sect:GSy}.
\item \texttt{iGS}, which has been introduced in Section~\ref{sect:iGS}.
\end{enumerate}

All six algorithms built an RR model from the labeled samples, which minimizes the following regularized loss function:
\begin{align}
l(\lambda,\boldsymbol{\beta})=\sum_{m=1}^K (y_m-\boldsymbol{\beta}^T\mathbf{x}_m)^2+\lambda|\boldsymbol{\beta}|^2 \label{eq:RR}
\end{align}
where $\boldsymbol{\beta}$ contains the regression coefficients, and $\lambda=0.01$ was used in our study. We used RR instead of ordinary least squares linear regression because the number of labeled samples is very small, so RR, with regularization on the regression coefficients, generally results in better generalization performance than the ordinary linear regression.

\subsection{Evaluation Process} \label{sect:process}

The evaluation process was similar to those used in our previous research on pool-based ALR \cite{drwuEBMAL2016,drwuSAL2018}. For each dataset, we first randomly selected 80\% of the total samples as the pool\footnote{For a fixed pool, \texttt{GSx} gives a deterministic selection sequence because it does not involve randomness. So, we need to vary the pool in order to study its statistical property.}, initialized the first $K_0$ labeled samples ($K_0$ is the dimensionality of the input space) either randomly (for \texttt{BL}, \texttt{QBC} and \texttt{EMCM}) or by \texttt{GSx} (for \texttt{GSx}, \texttt{GSy} and \texttt{iGS}), identified one sample to label in each iteration by different algorithms, and built an RR model. The maximum number of samples to be labeled, $K$, was 20\% of the dataset size. For datasets too small or too large, we constrained $K\in[20,60]$.

To obtain statistically meaningful results, we ran this evaluation process 100 times for each dataset and each algorithm, each time with a randomly chosen 80\% population pool.

\subsection{Performance Measures} \label{sect:pm}

After each iteration of each algorithm, we computed the root mean squared error (RMSE) and correlation coefficient (CC) as the performance measures.

Because different algorithms selected different samples to label, the remaining unlabeled samples in the pool were different for each algorithm, so we cannot compare their performances based on the remaining unlabeled samples. Because in pool-based ALR the goal is to build a regression model to label all samples in the pool as accurately as possible, we computed the RMSE and CC using all samples in the pool, where the labels for the $K$ selected samples were their true labels, and the labels for the remaining $N-K$ unlabeled samples were the estimates from the regression model.

Note that although both RMSE and CC were used as our performance measures, we should consider the RMSE as the \emph{primary} one, because it was directly optimized in the objective function of the regression model [see (\ref{eq:RR})]. Generally as the RMSE decreases, the CC should increase, but not always. In other words, we expect that an ALR approach that gives a small RMSE should also have a large CC, but this is not always true. So, the CC can only be viewed as a \emph{secondary} performance measure.

\subsection{Experimental Results}

\begin{figure*}[!h]\centering
\subfigure[]{\label{fig:ConcreteCS}     \includegraphics[width=.32\linewidth,clip]{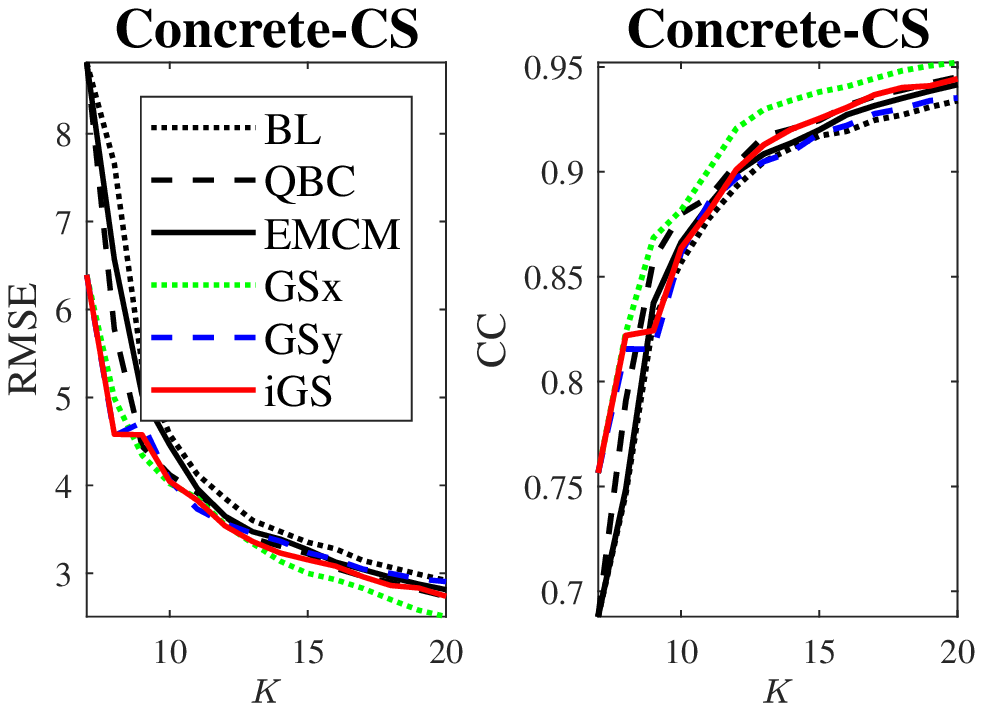}}
\subfigure[]{\label{fig:ConcreteFlow}     \includegraphics[width=.32\linewidth,clip]{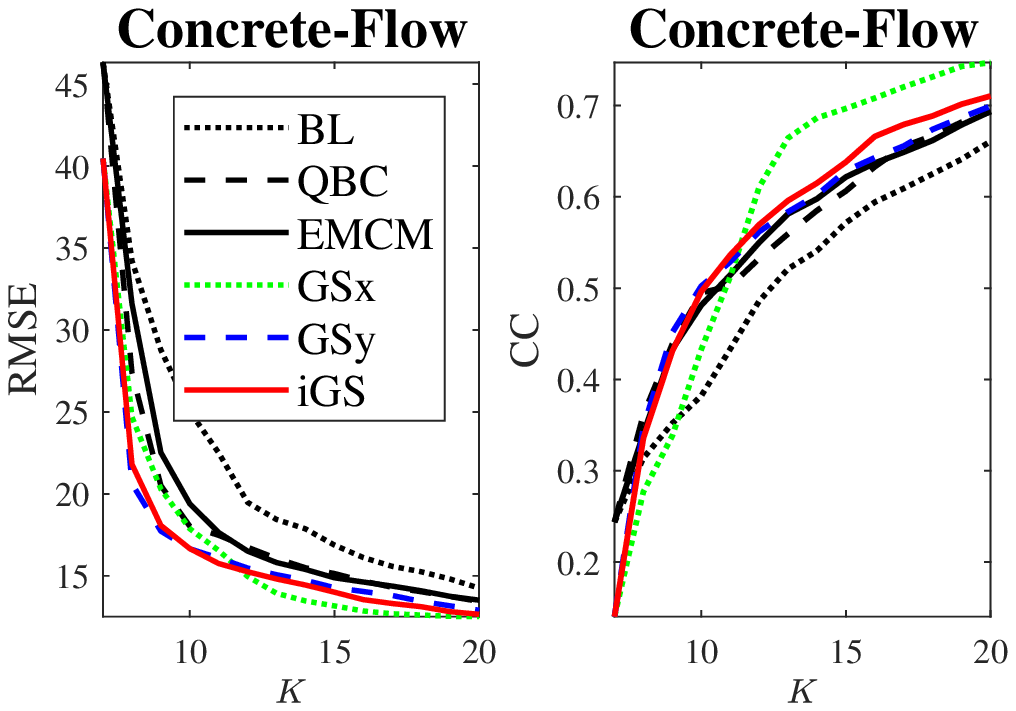}}
\subfigure[]{\label{fig:ConcreteSlump}     \includegraphics[width=.32\linewidth,clip]{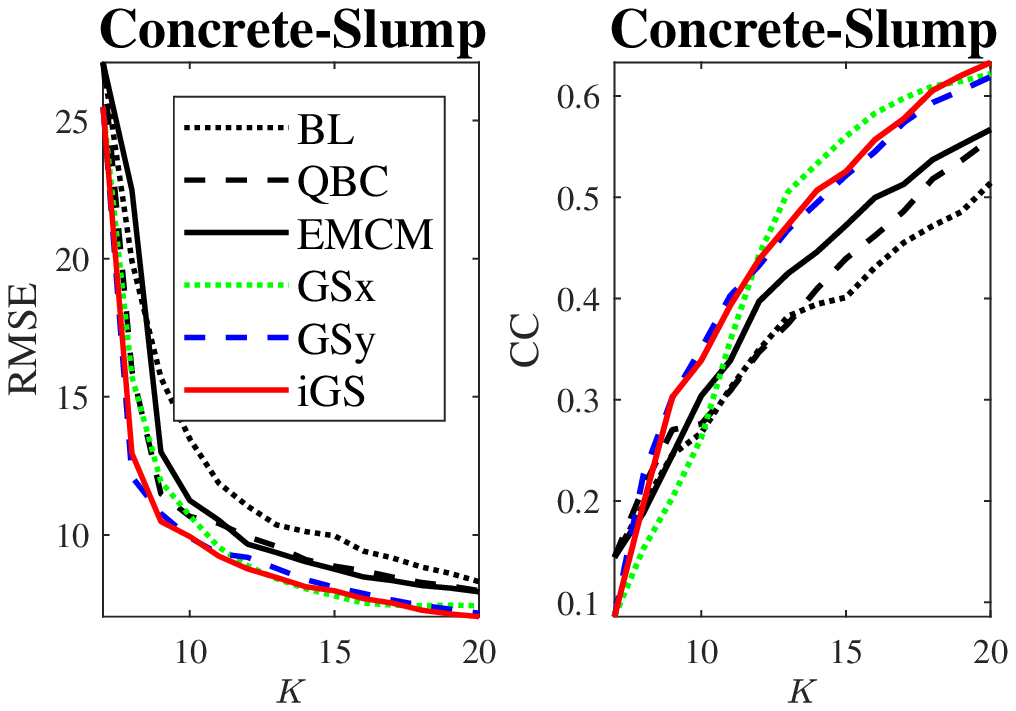}}
\subfigure[]{\label{fig:Yacht}     \includegraphics[width=.32\linewidth,clip]{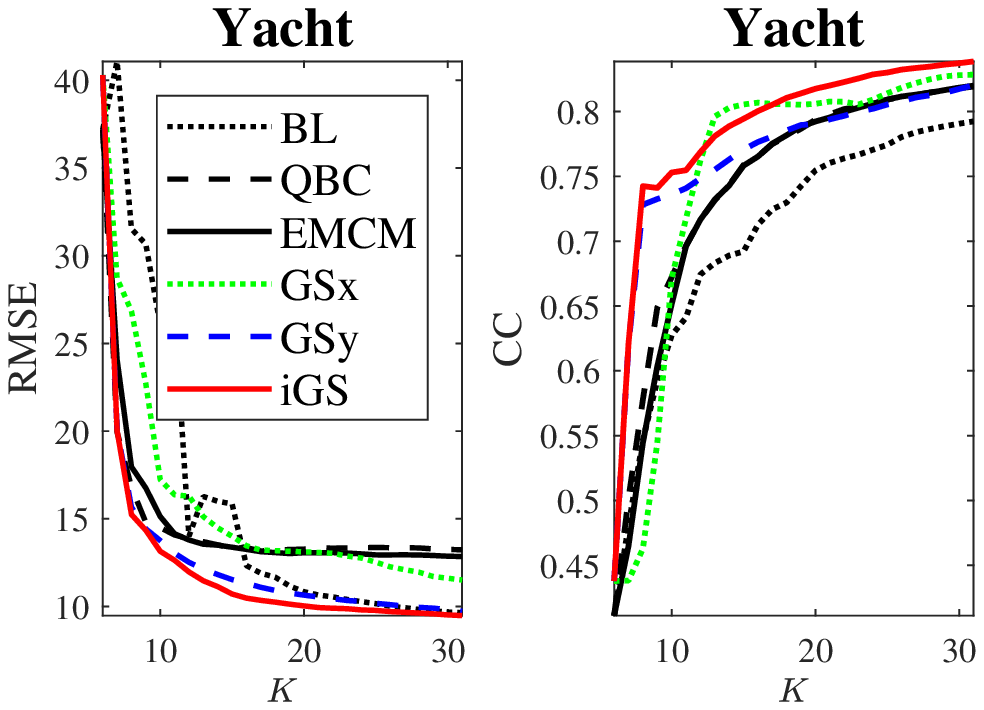}}
\subfigure[]{\label{fig:autoMPG}     \includegraphics[width=.32\linewidth,clip]{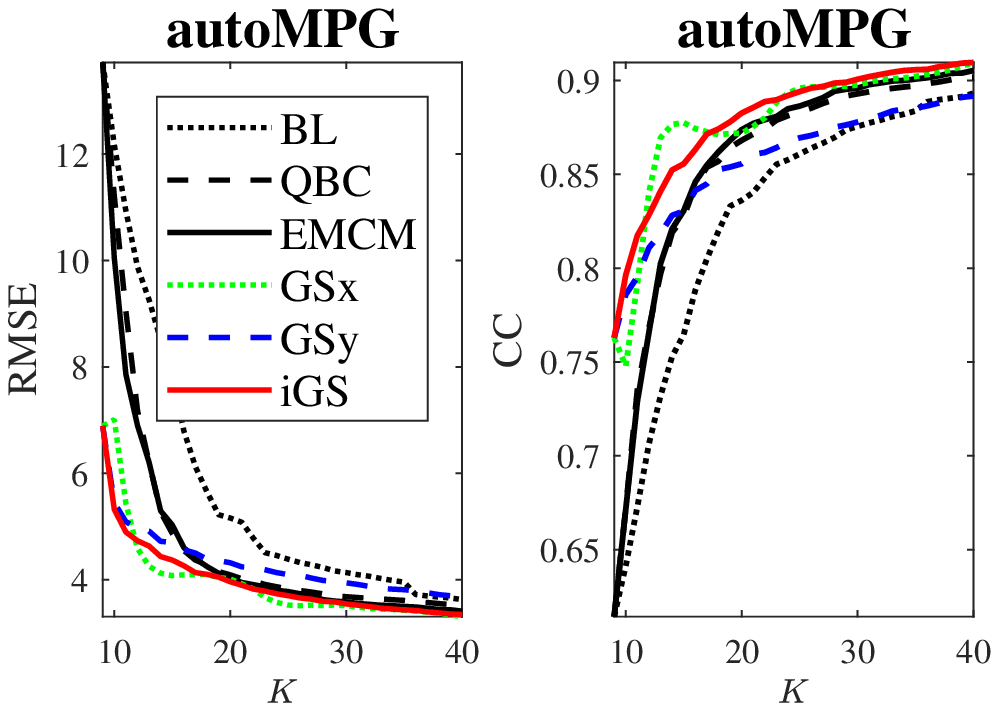}}
\subfigure[]{\label{fig:NO2}     \includegraphics[width=.32\linewidth,clip]{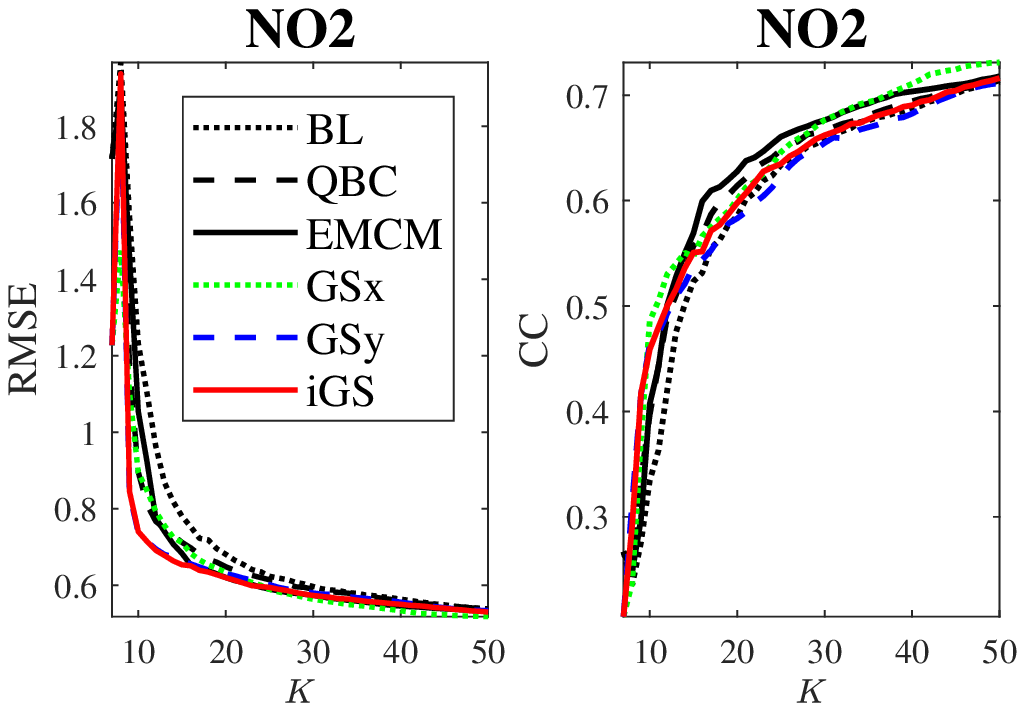}}
\subfigure[]{\label{fig:PM10}     \includegraphics[width=.32\linewidth,clip]{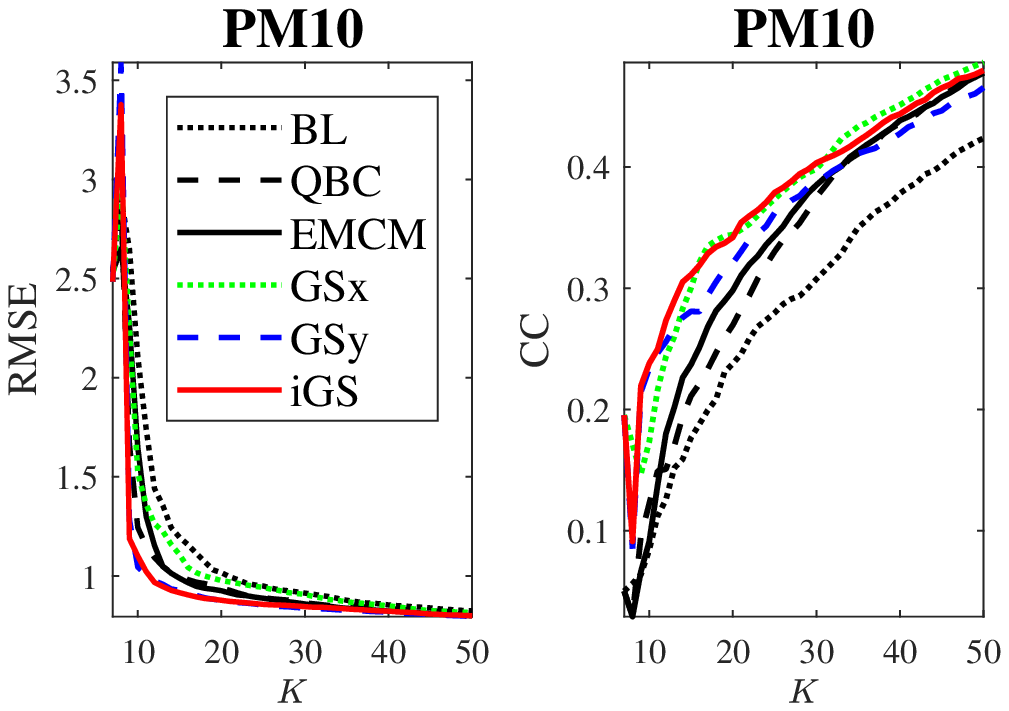}}
\subfigure[]{\label{fig:Housing}     \includegraphics[width=.32\linewidth,clip]{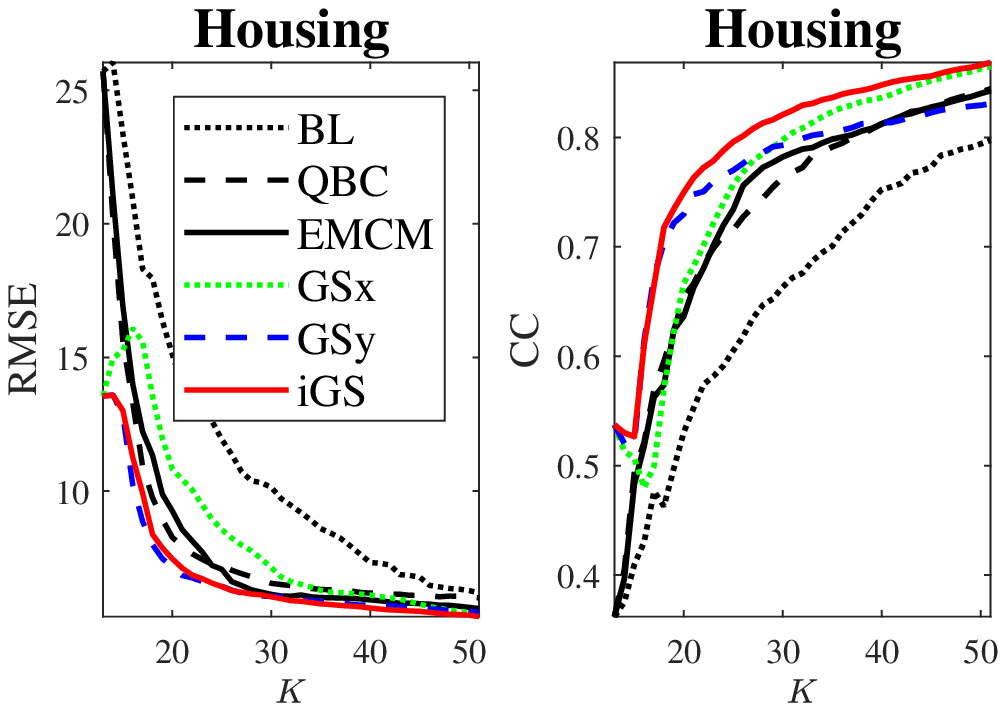}}
\subfigure[]{\label{fig:CPS}     \includegraphics[width=.32\linewidth,clip]{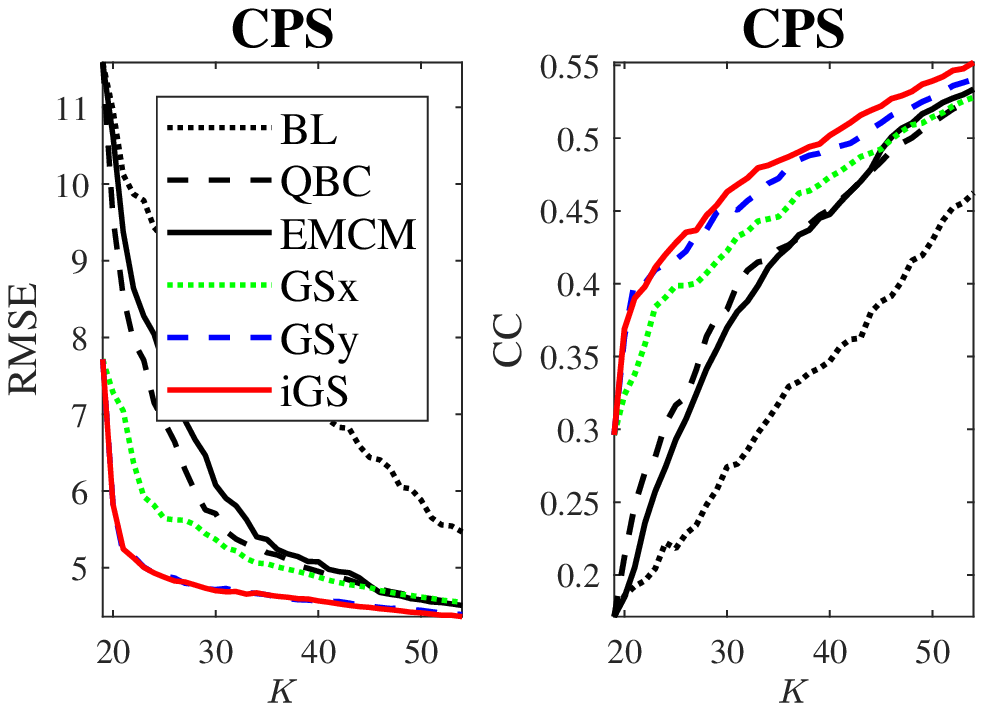}}
\subfigure[]{\label{fig:Concrete}     \includegraphics[width=.32\linewidth,clip]{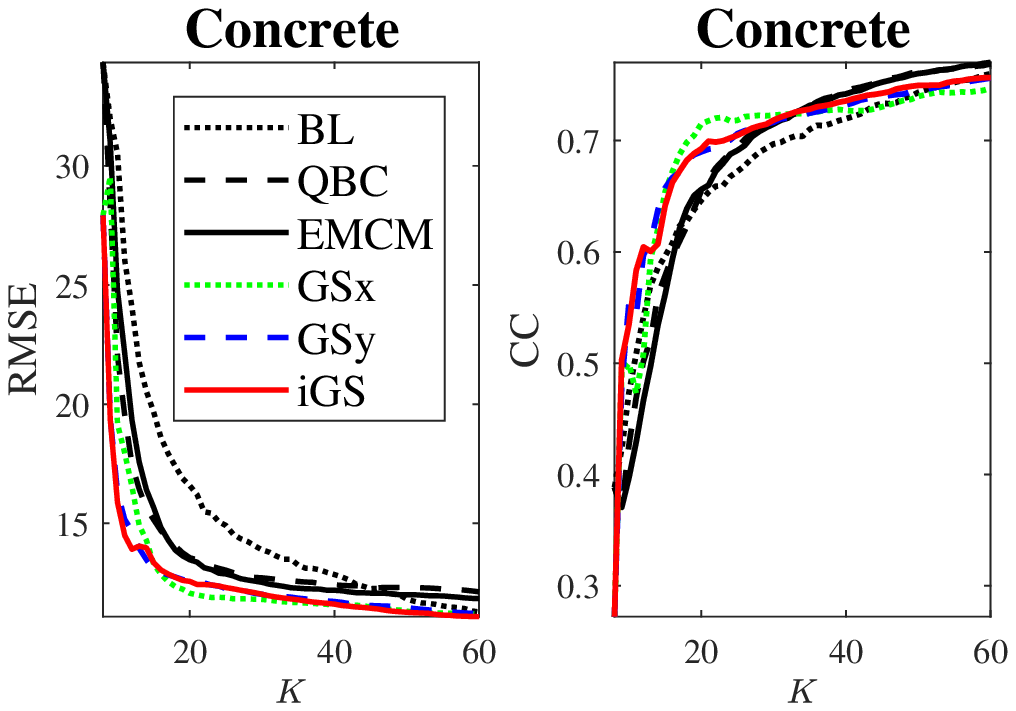}}
\subfigure[]{\label{fig:Wine-red}     \includegraphics[width=.32\linewidth,clip]{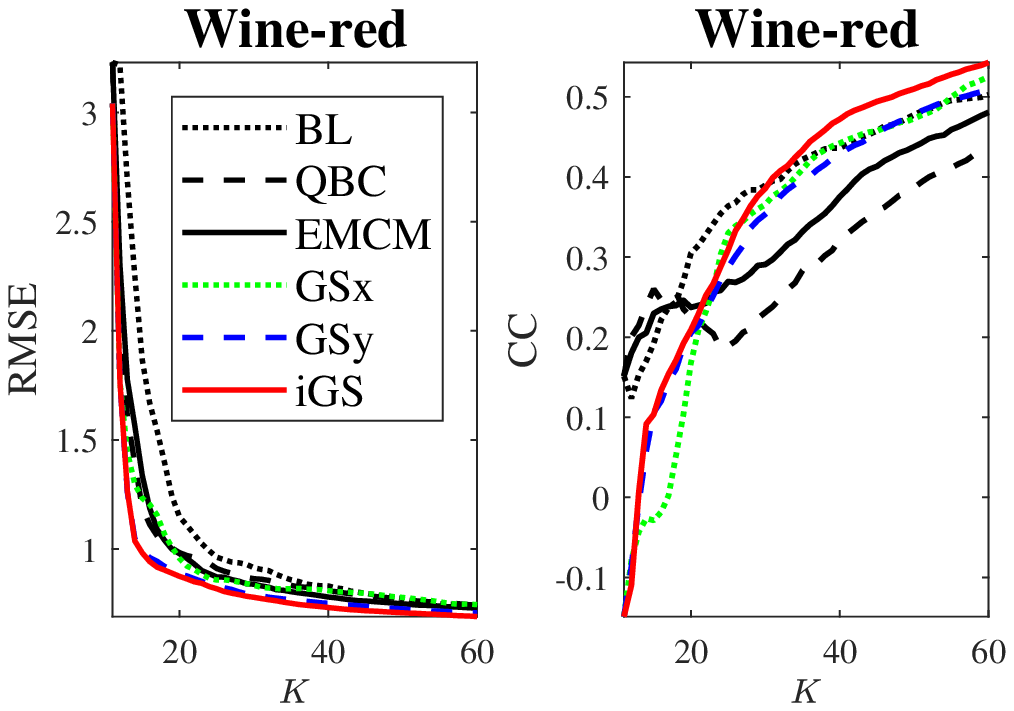}}
\subfigure[]{\label{fig:Wine-white}     \includegraphics[width=.32\linewidth,clip]{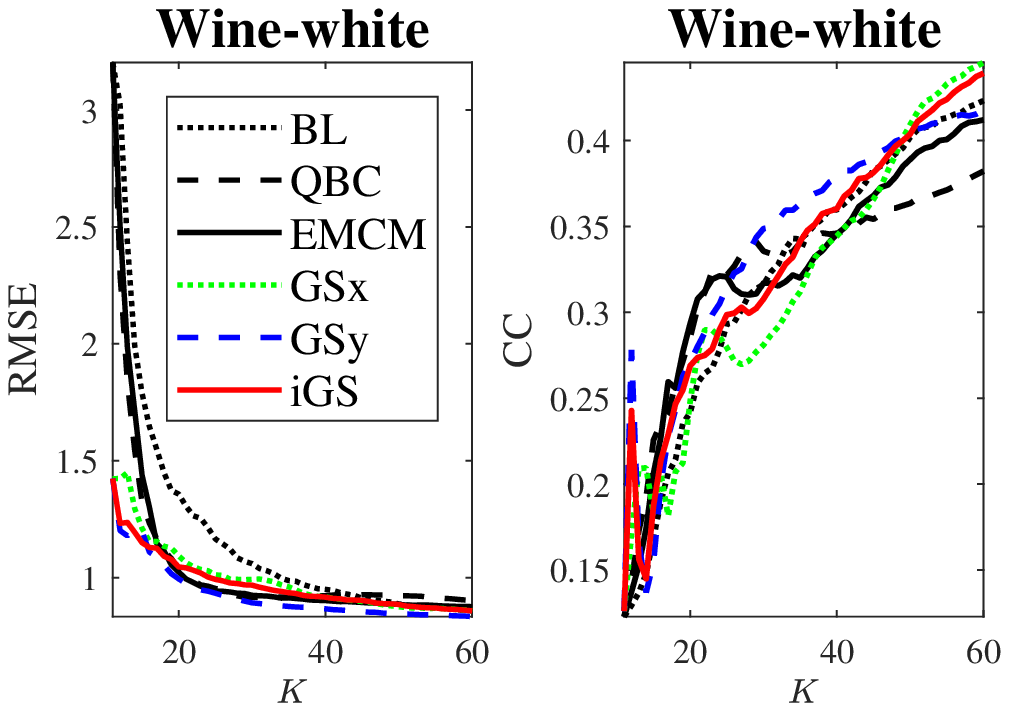}}
\caption{Performances of the six algorithms on the 12 datasets, averaged over 100 runs. (a) Concrete-CS; (b) Concrete-Flow; (c) Concrete-Slump; (d) Yacht; (e) autoMPG; (f) NO2; (g) PM10; (h) Housing; (i) CPS; (j) Concrete; (k) Wine-red; (l) Wine-white.} \label{fig:results12}
\end{figure*}

The RMSEs and CCs for the six algorithms on the 12 datasets, averaged over 100 runs, are shown in Figure~\ref{fig:results12}. Generally as $K$ increased, all six algorithms achieved better performance (smaller RMSE and larger CC), which is intuitive, because more labeled training samples generally result in a more reliable RR model. \texttt{iGS} achieved the smallest RMSE and largest CC on most datasets.

\begin{figure}[!h]\centering
\subfigure[]{\label{fig:AUC-RMSE-UCI}     \includegraphics[width=\linewidth,clip]{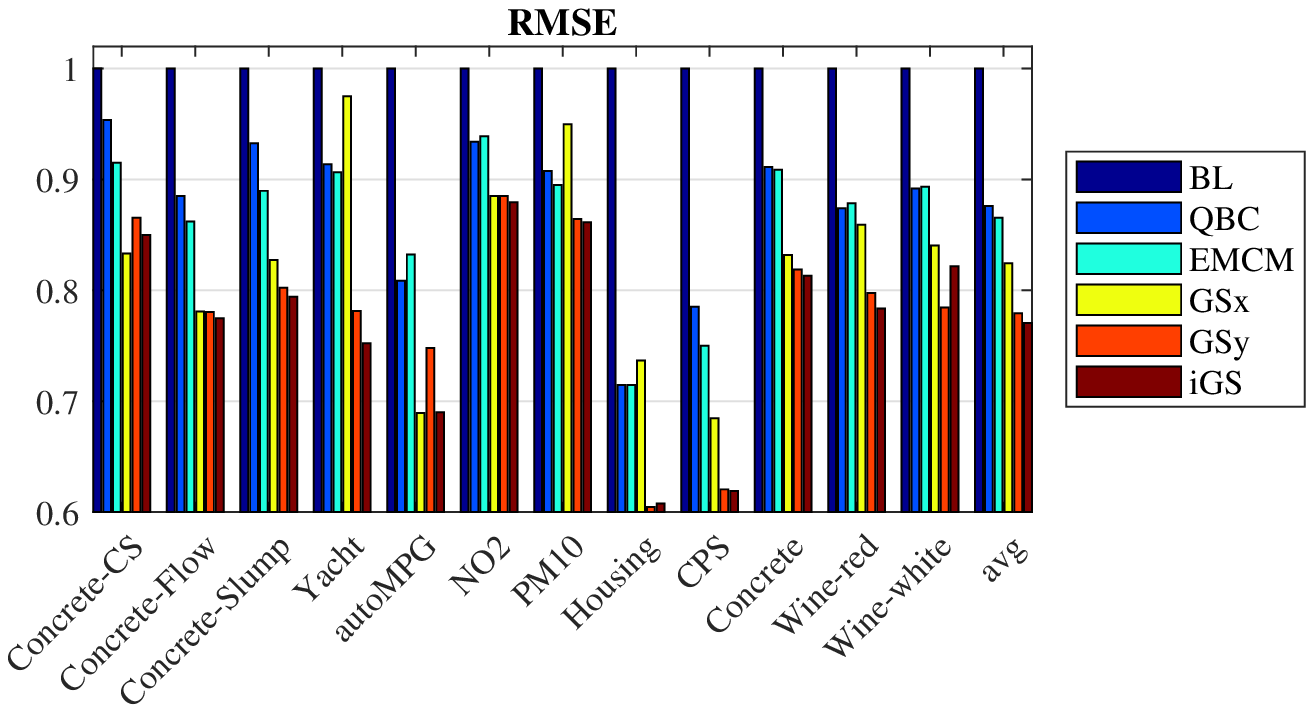}}
\subfigure[]{\label{fig:AUC-CC-UCI}     \includegraphics[width=\linewidth,clip]{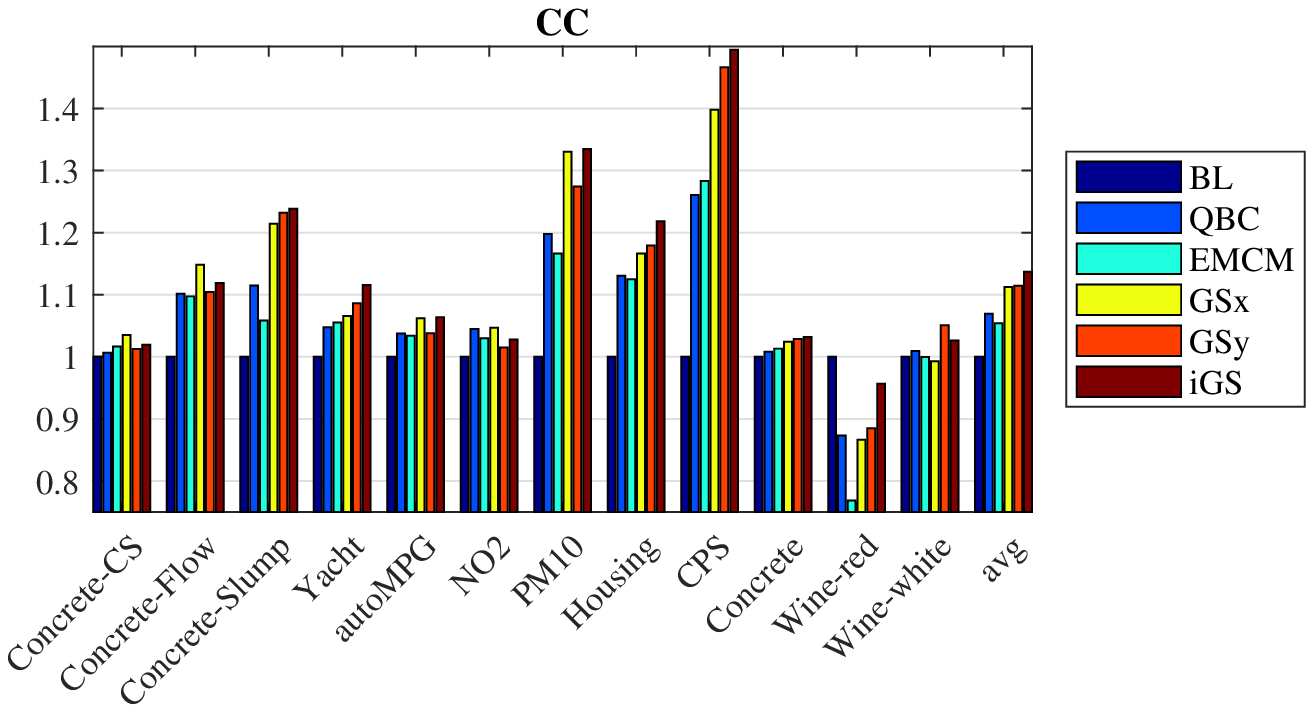}}
\caption{Normalized AUCs of the six algorithms on the 12 datasets. (a) RMSE; (b) CC.} \label{fig:AUC}
\end{figure}

To see the forest for the trees, we also define an aggregated performance measure called the area under the curve (AUC) for the average RMSE and the average CC on each of the 12 datasets in Figure~\ref{fig:results12}. The AUCs for the RMSEs are shown in Figure~\ref{fig:AUC-RMSE-UCI}, where for each dataset, we used the AUC of BL to normalize the AUCs of the other five algorithms, so the AUC of \texttt{BL} was always 1. For the RMSE, a smaller AUC indicates a better performance. Similarly, we also show the normalized AUCs of the CCs in Figure~\ref{fig:AUC-CC-UCI}, where a larger AUC indicates a better performance. Figure~\ref{fig:AUC} shows that:
\begin{enumerate}
\item Each of \texttt{QBC}, \texttt{EMCM}, \texttt{GSx}, \texttt{GSy} and \texttt{iGS} achieved smaller RMSEs than \texttt{BL} on all 12 datasets, and larger CCs than \texttt{BL} on at least 10 of the 12 datasets, suggesting that these five ALR approaches were all effective.
\item On average the performances of the six algorithms, from the best to the worst, was \texttt{iGS} $>$ \texttt{GSy} $>$ \texttt{GSx} $>$ \texttt{EMCM} $\approx$ \texttt{QBC} $>$ \texttt{BL}. \texttt{iGS} combines the advantages of \texttt{GSx} and \texttt{GSy}, and outperformed both of them.
\end{enumerate}
These observations were also confirmed by Table~\ref{tab:ranksUCI}, which shows the detailed ranks of the six approaches on the 12 datasets, according to the AUCs.

\begin{table}[!h] \centering \setlength{\tabcolsep}{1.5mm}
\caption{Ranks of the six approaches on the 12 datasets.}   \label{tab:ranksUCI}
\begin{tabular}{c|l|cccccc}   \hline
   &Dataset & \texttt{BL}  & \texttt{QBC} & \texttt{EMCM} & \texttt{GSx} & \texttt{GSy} & \texttt{iGS} \\ \hline
   &Concrete-CS        &6&5&4&1&3&2 \\
   &Concrete-Flow      &6&5&4&3&2&1 \\
   &Concrete-Slump     &6&5&4&3&2&1 \\
   &Yacht              &6&4&3&5&2&1 \\
   &autoMPG            &6&4&5&1&3&2 \\
   &NO2                &6&4&5&2&3&1 \\
RMSE&PM10              &6&4&3&5&2&1 \\
   &Housing            &6&4&3&5&1&2 \\
   &CPS                &6&5&4&3&2&1 \\
   &Concrete           &6&5&4&3&2&1 \\
   &Wine-red           &6&4&5&3&2&1 \\
   &Wine-white         &6&4&5&3&1&2  \\
   &\textbf{Average}    &\textbf{6}&\textbf{5}&\textbf{4}&\textbf{3}&\textbf{2}&\textbf{1}\\  \hline
   &Concrete-CS        &6&5&3&1&4&2\\
   &Concrete-Flow      &6&4&5&1&3&2\\
   &Concrete-Slump     &6&4&5&3&2&1\\
   &Yacht              &6&5&4&3&2&1\\
   &autoMPG            &6&4&5&2&3&1\\
   &NO2                &6&2&3&1&5&4\\
CC &PM10               &6&4&5&2&3&1\\
   &Housing            &6&4&5&3&2&1\\
   &CPS                &6&5&4&3&2&1\\
   &Concrete           &6&5&4&3&2&1\\
   &Wine-red           &1&4&6&5&3&2\\
   &Wine-white         &4&3&5&6&1&2  \\
   &\textbf{Average}   &\textbf{6}&\textbf{4}&\textbf{5}&\textbf{3}&\textbf{2}&\textbf{1}\\ \hline
\end{tabular}
\end{table}
\renewcommand{\baselinestretch}{1.5}

\subsection{Statistical Analysis} \label{sect:SA}

To determine if the differences between different pairs of algorithms were statistically significant, we also performed non-parametric multiple comparison tests on the AUCs using Dunn's procedure \cite{Dunn1961,Dunn1964}, with a $p$-value correction using the False Discovery Rate method \cite{Benjamini1995}. The $p$-values for the AUCs of RMSEs and CCs are shown in Table~\ref{tab:DunnUCI}, where the statistically significant ones are marked in bold. Table~\ref{tab:DunnUCI} shows that:
\begin{enumerate}
\item All five ALR approaches had statistically significantly better RMSEs and CCs than \texttt{BL}, suggesting again that they were effective.
\item Among the three existing ALR approaches, \texttt{GSx} had statistically significantly better RMSE and CC than both \texttt{QBC} and \texttt{EMCM}.
\item \texttt{GSy} had statistically significantly better RMSE and CC than both \texttt{QBC} and \texttt{EMCM}, and also statistically significantly better RMSE than \texttt{GSx}.
\item \texttt{iGS} had statistically significantly better RMSE than all other approaches except \texttt{GSy}, and also statistically significantly better CC than all other approaches except \texttt{GSx}.
\end{enumerate}
These observations verified the effectiveness of the two proposed approaches, particularly \texttt{iGS}.

\begin{table}[!h] \centering \setlength{\tabcolsep}{2mm}
\caption{$p$-values of non-parametric multiple comparisons on the AUCs of RMSEs and CCs on the 12 UCI and CMU StatLib datasets.}   \label{tab:DunnUCI}
\begin{tabular}{c|l|ccccc}   \hline
&   & \texttt{BL} &           \texttt{QBC} &     \texttt{EMCM}  & \texttt{GSx} &  \texttt{GSy}   \\ \hline
&\texttt{QBC} & \textbf{.0000} & & & &\\
&\texttt{EMCM} & \textbf{.0000} & .1096 & & & \\
RMSE&\texttt{GSx} & \textbf{.0000} & \textbf{.0000} & \textbf{.0000}& \\
&\texttt{GSy} & \textbf{.0000} & \textbf{.0000} & \textbf{.0000} &\textbf{.0000} & \\
&\texttt{iGS} & \textbf{.0000} & \textbf{.0000} & \textbf{.0000} & \textbf{.0002} &.0715 \\   \hline
&\texttt{QBC} & \textbf{.0000} & & & &\\
&\texttt{EMCM} & \textbf{.0000} & .0805 & & & \\
CC&\texttt{GSx} & \textbf{.0000} & \textbf{.0000} & \textbf{.0000}& \\
&\texttt{GSy} & \textbf{.0000} & \textbf{.0015} & \textbf{.0000} &.1504 & \\
&\texttt{iGS} & \textbf{.0000} & \textbf{.0000} & \textbf{.0000} & .0290 &\textbf{.0017} \\   \hline
\end{tabular}
\end{table}

\section{Experiments on EEG-Based Driver Drowsiness Estimation} \label{sect:Driving}

Experiments on offline EEG-based driver drowsiness estimation are performed in this section to further demonstrate the performances of \texttt{GSy} and \texttt{iGS}.

\subsection{Experiment Setup}

The experiment and data used in \cite{drwuEBMAL2016} was again used in this paper. Sixteen healthy subjects with normal/corrected-to-normal vision participated in a sustained-attention driving experiment \cite{Chuang2012,Chuang2014}, consisting of a real vehicle mounted on a motion platform with 6 degrees of freedom immersed in a 360$^\circ$ virtual-reality scene, simulating monotonous driving at 100 km/h on a straight and empty highway. Each experiment was conducted for about 60-90 minutes in the afternoon when the circadian rhythm of sleepiness reached its peak. Random lane-departure disturbances were applied every 5-10 seconds, and participants needed to steer the vehicle to compensate for them as quickly as possible. The response time was recorded and later converted to a drowsiness index. Participants' scalp EEG signals were also recorded using a 500Hz 32-channel Neuroscan system (30-channel EEGs plus 2-channel earlobes).

The Institutional Review Board of the Taipei Veterans General Hospital approved the experimental protocol.

\subsection{Preprocessing and Feature Extraction}

The preprocessing and feature extraction procedures were almost identical to those in our recent research \cite{drwuEBMAL2016}.

The 16 subjects had different lengths of experiment, because the disturbances were presented randomly every 5-10 seconds. To ensure fair comparison, we used only the first 3,600 seconds data for each subject. Data from one subject was not recorded correctly, so we used only 15 subjects.

We defined a function \cite{drwuaBCI2015} to map the response time $\tau$ to a drowsiness index $y\in[0, 1]$:
\begin{align}
y=\max\left\{0,\,\frac{1-e^{-(\tau-\tau_0)}}{1+e^{-(\tau-\tau_0)}}\right\} \label{eq:y}
\end{align}
$\tau_0=1$ was used in this paper. The drowsiness indices were then smoothed using a 90-second square moving-average window to reduce variations.

EEGLAB \cite{Delorme2004} was used for EEG signal preprocessing. After 1-50 Hz band-pass filtering, the EEG data were downsampled from 500 Hz to 250 Hz and re-referenced to averaged earlobes.

Our goal was to predict the drowsiness index for each subject every 10 seconds (called a sample point in this paper). All 30 EEG channels were used in feature extraction. We epoched 30-second EEG signals right before each sample point, and computed the average power spectral density (PSD) in the theta band (4-7.5 Hz) for each channel using Welch's method \cite{Welch1967}. Next, we converted the 30 theta band powers to dBs. To remove noises or bad channel readings, we removed channels whose maximum dBs were larger than 20. We then normalized the dBs of each remaining channel to mean zero and standard deviation one, and extracted 10 leading principal components. The projections of the dBs onto these principal components were then normalized to $[0, 1]$ and used as our features.

\subsection{Experimental Results}

The six algorithms introduced in Section~\ref{sect:algs} were again compared in this experiment. The evaluation process was the same as that in Section~\ref{sect:process}.

\begin{figure*}[!h]\centering
\subfigure[]{    \includegraphics[width=.32\linewidth,clip]{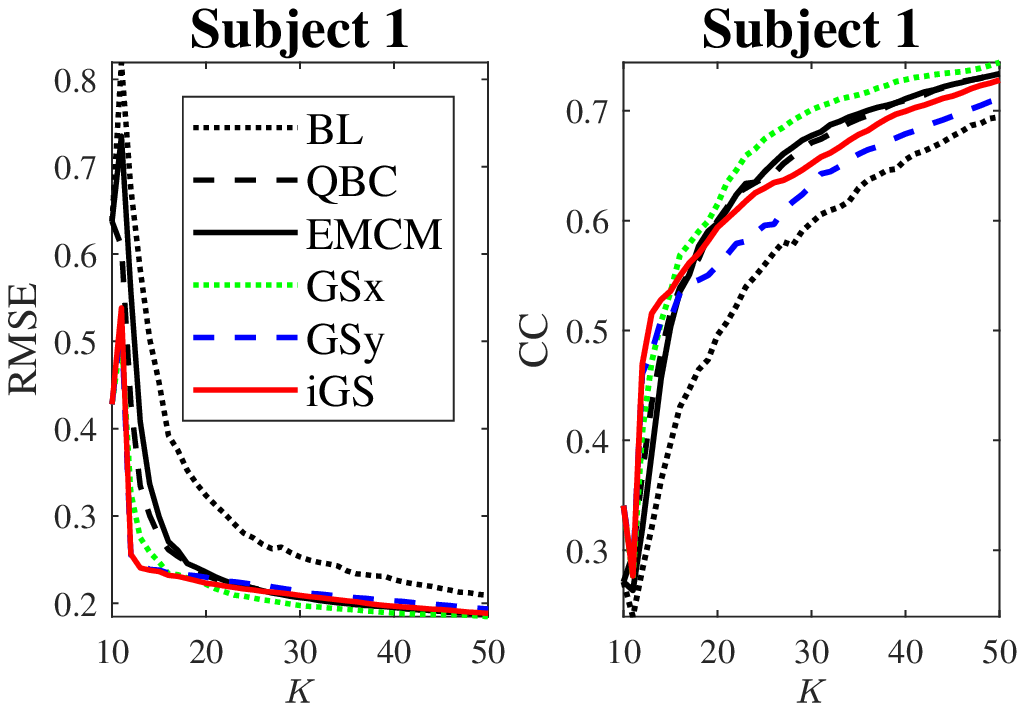}}
\subfigure[]{    \includegraphics[width=.32\linewidth,clip]{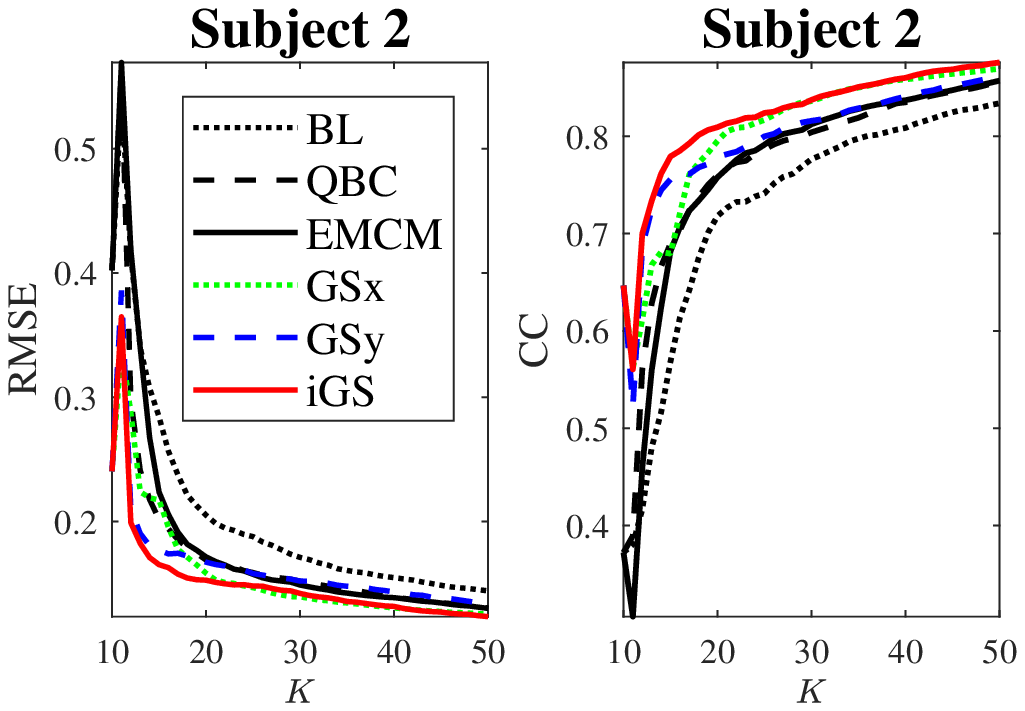}}
\subfigure[]{     \includegraphics[width=.32\linewidth,clip]{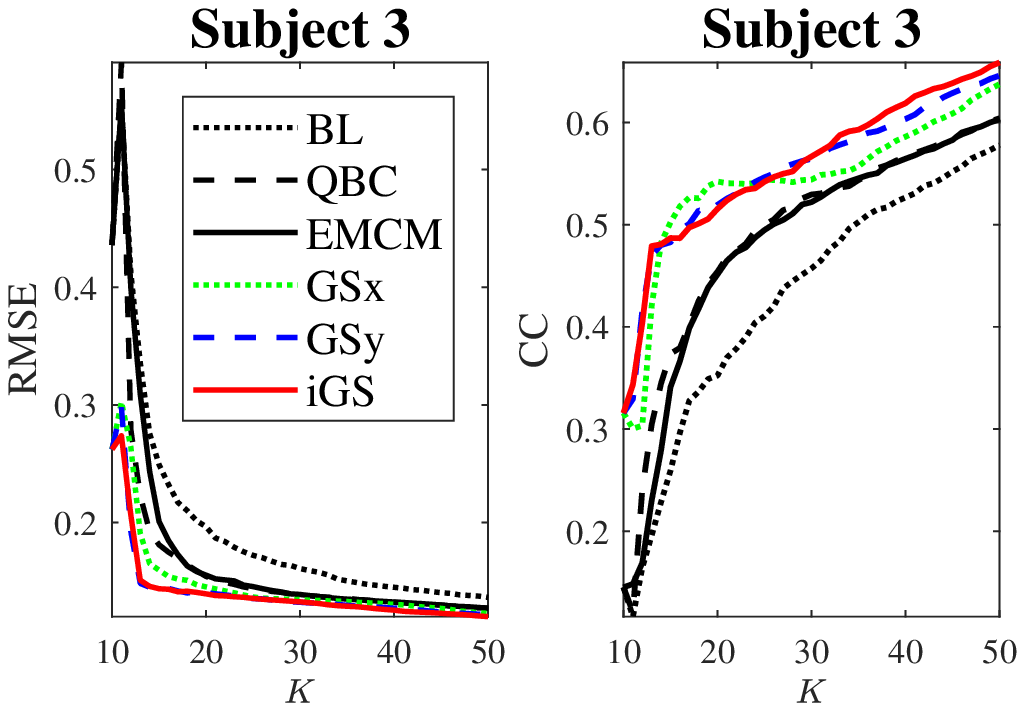}}
\subfigure[]{     \includegraphics[width=.32\linewidth,clip]{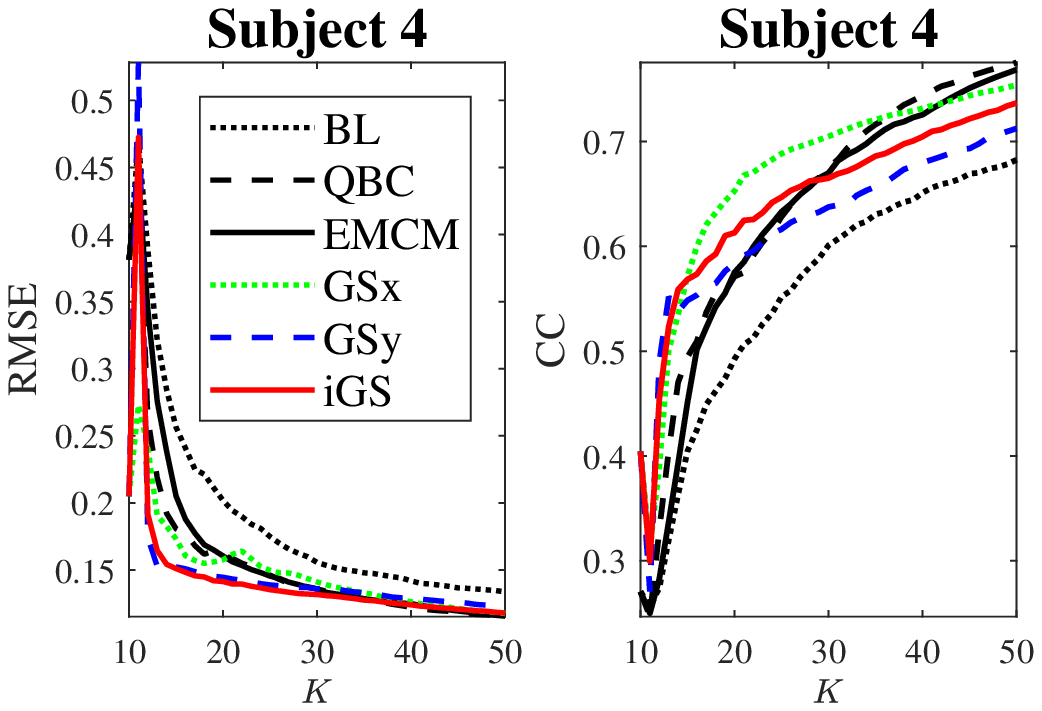}}
\subfigure[]{     \includegraphics[width=.32\linewidth,clip]{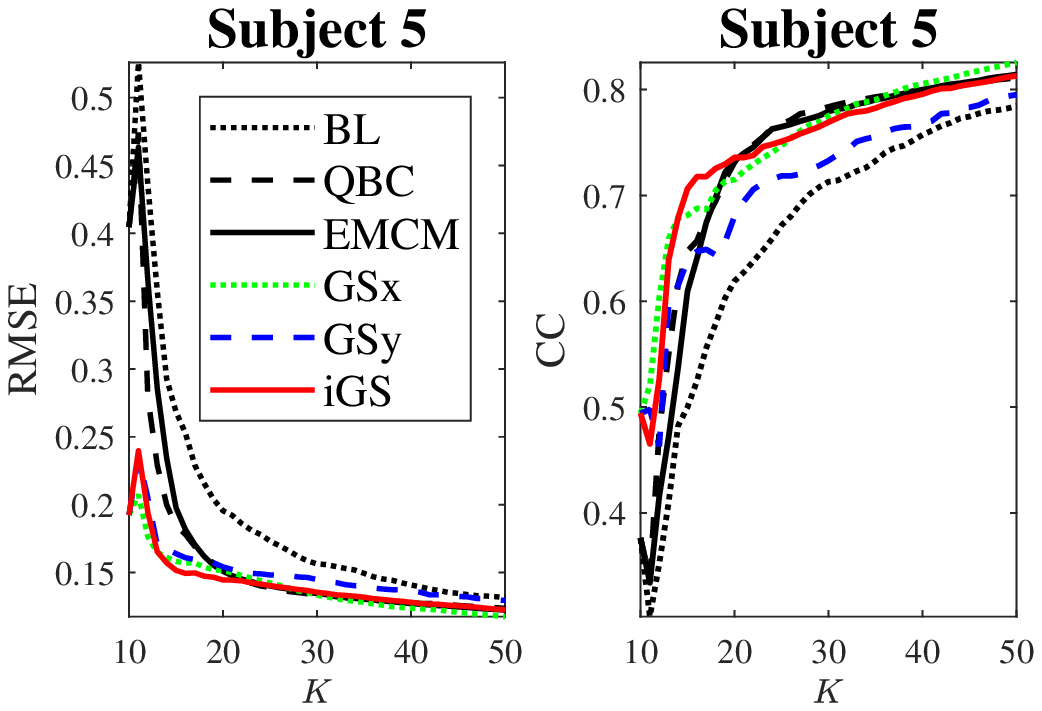}}
\subfigure[]{     \includegraphics[width=.32\linewidth,clip]{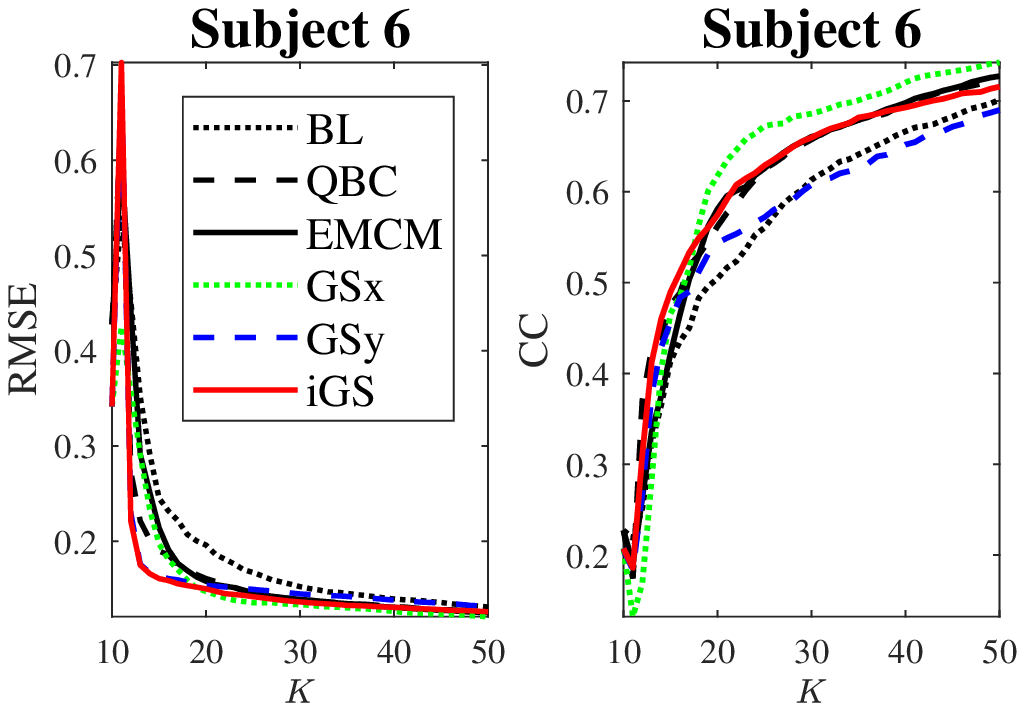}}
\subfigure[]{     \includegraphics[width=.32\linewidth,clip]{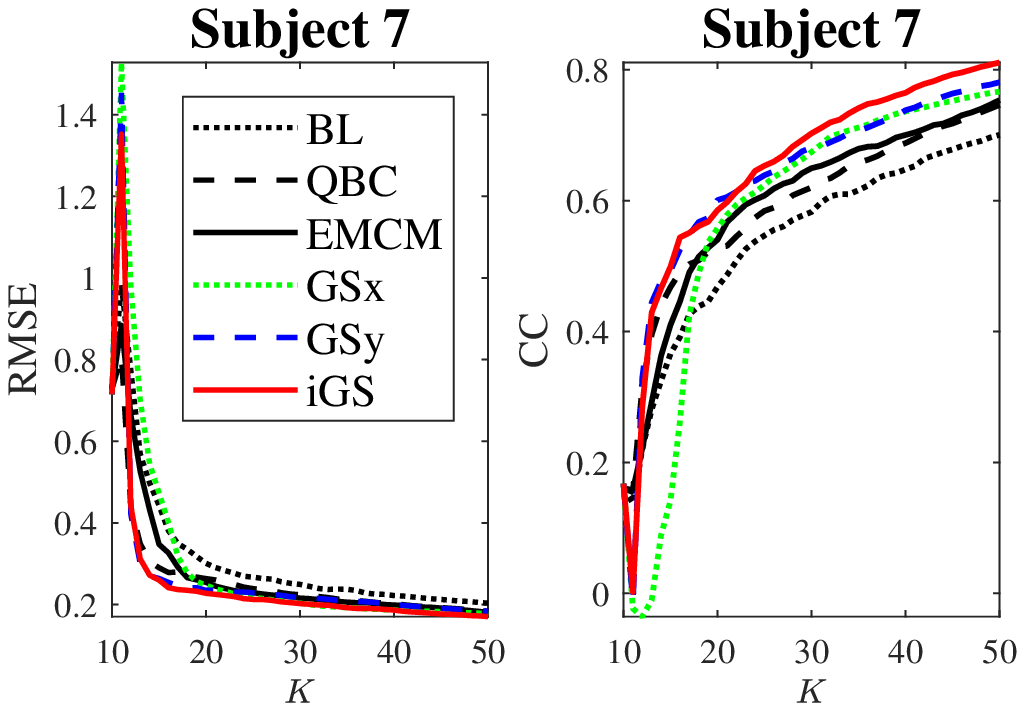}}
\subfigure[]{     \includegraphics[width=.32\linewidth,clip]{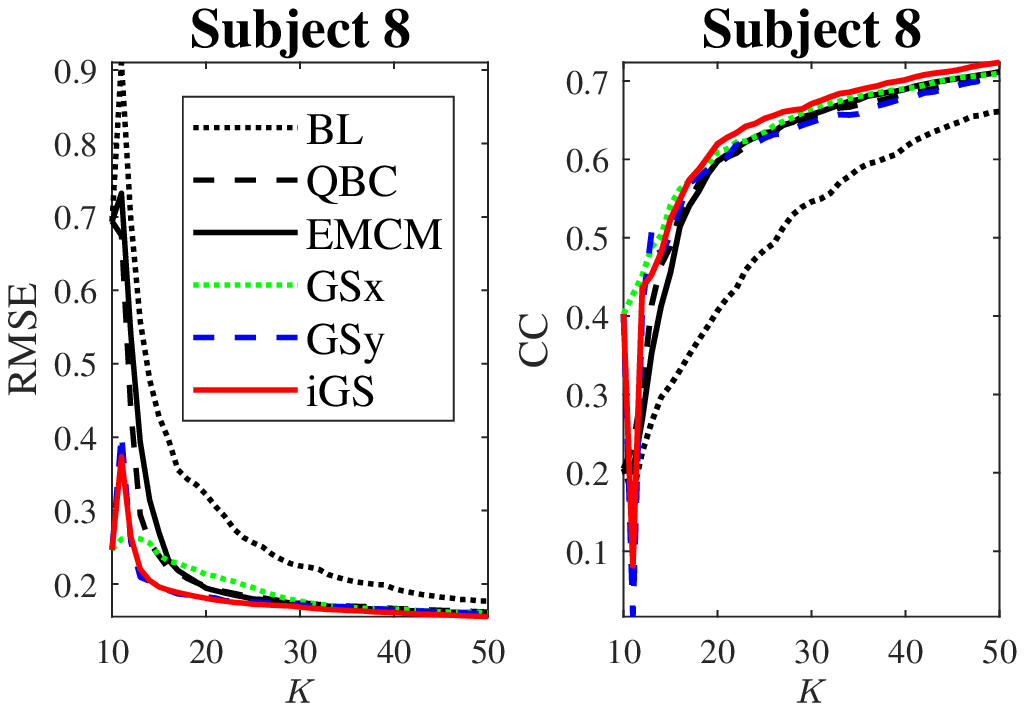}}
\subfigure[]{     \includegraphics[width=.32\linewidth,clip]{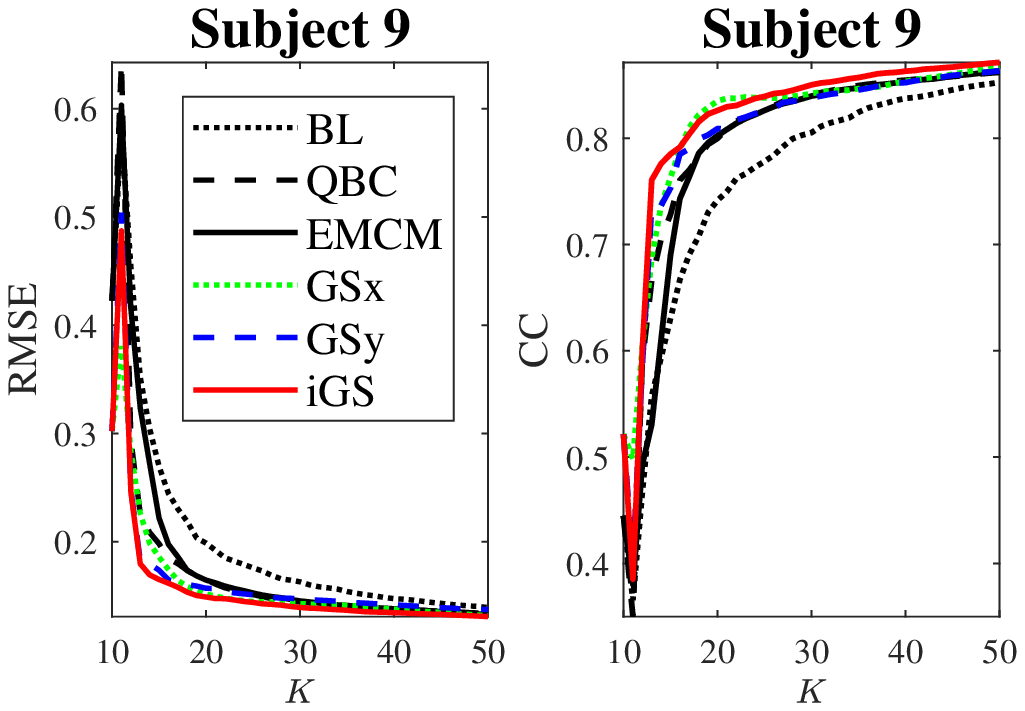}}
\subfigure[]{     \includegraphics[width=.32\linewidth,clip]{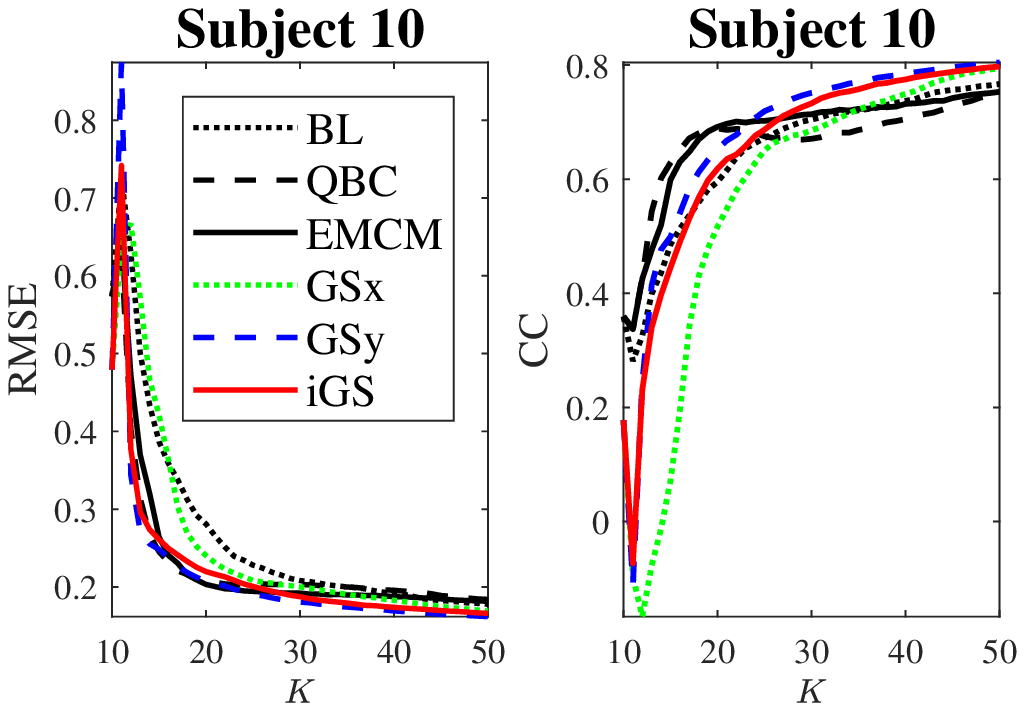}}
\subfigure[]{     \includegraphics[width=.32\linewidth,clip]{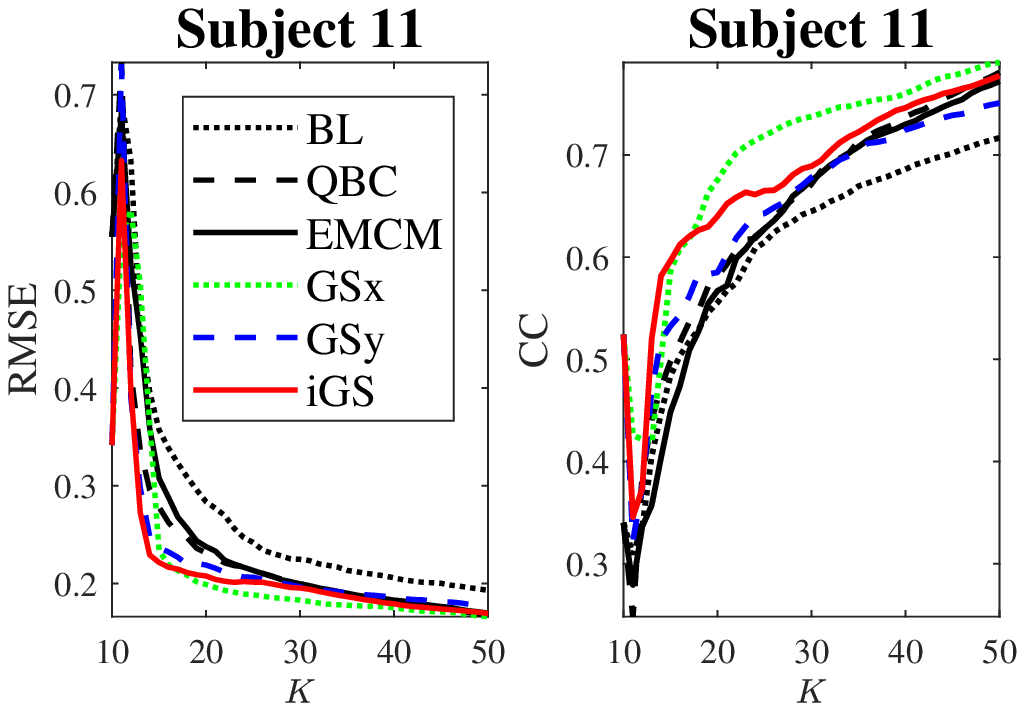}}
\subfigure[]{     \includegraphics[width=.32\linewidth,clip]{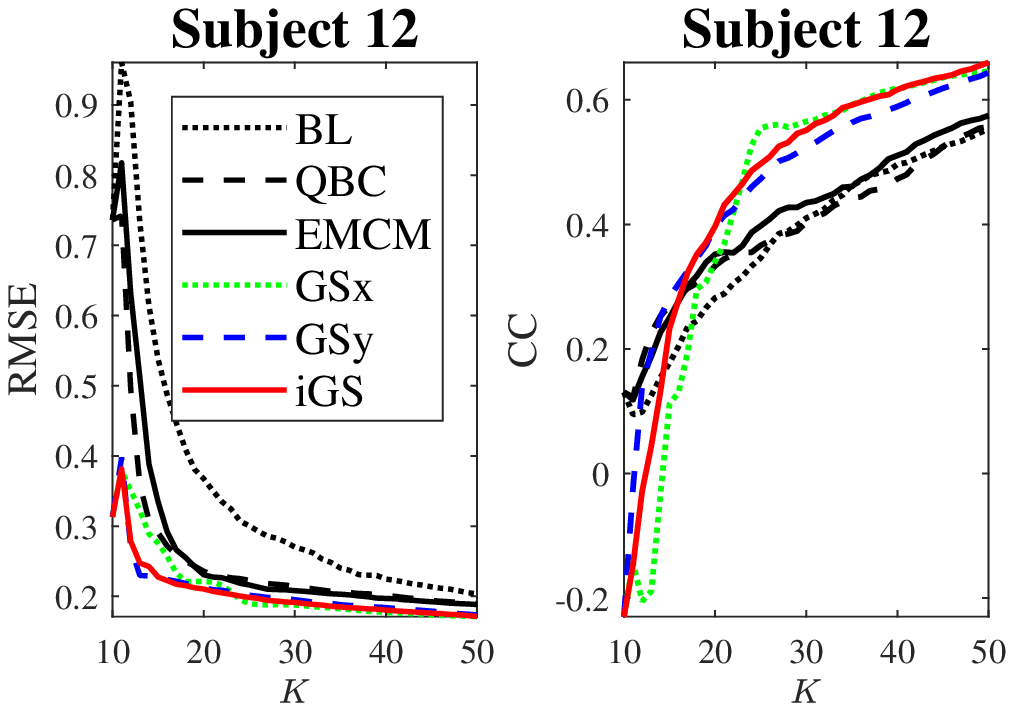}}
\subfigure[]{     \includegraphics[width=.32\linewidth,clip]{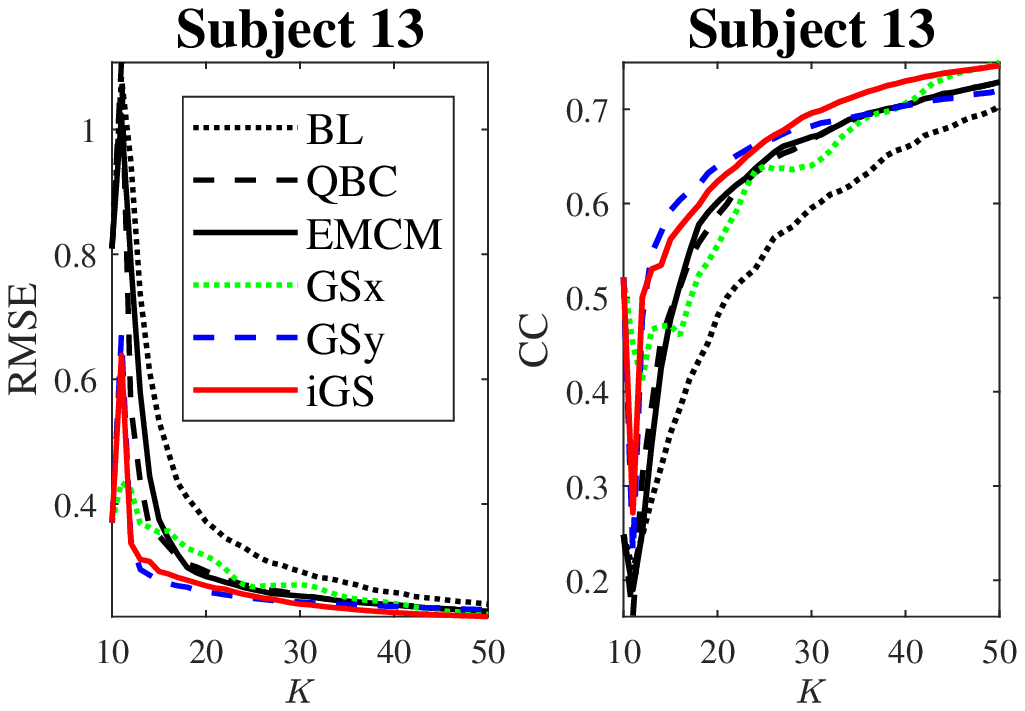}}
\subfigure[]{     \includegraphics[width=.32\linewidth,clip]{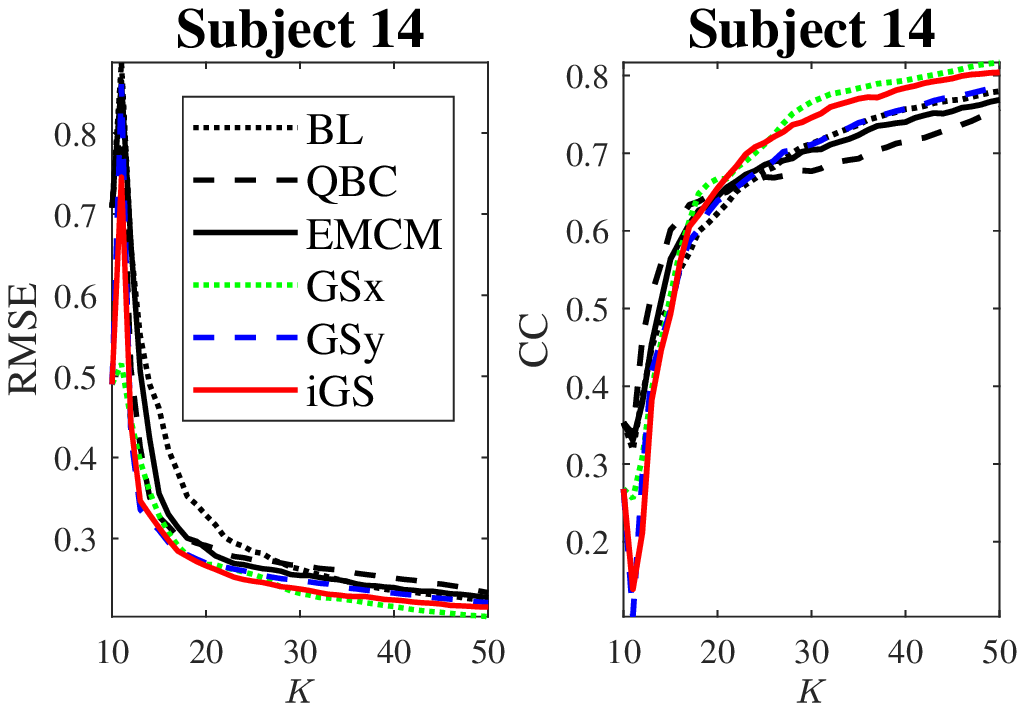}}
\subfigure[]{     \includegraphics[width=.32\linewidth,clip]{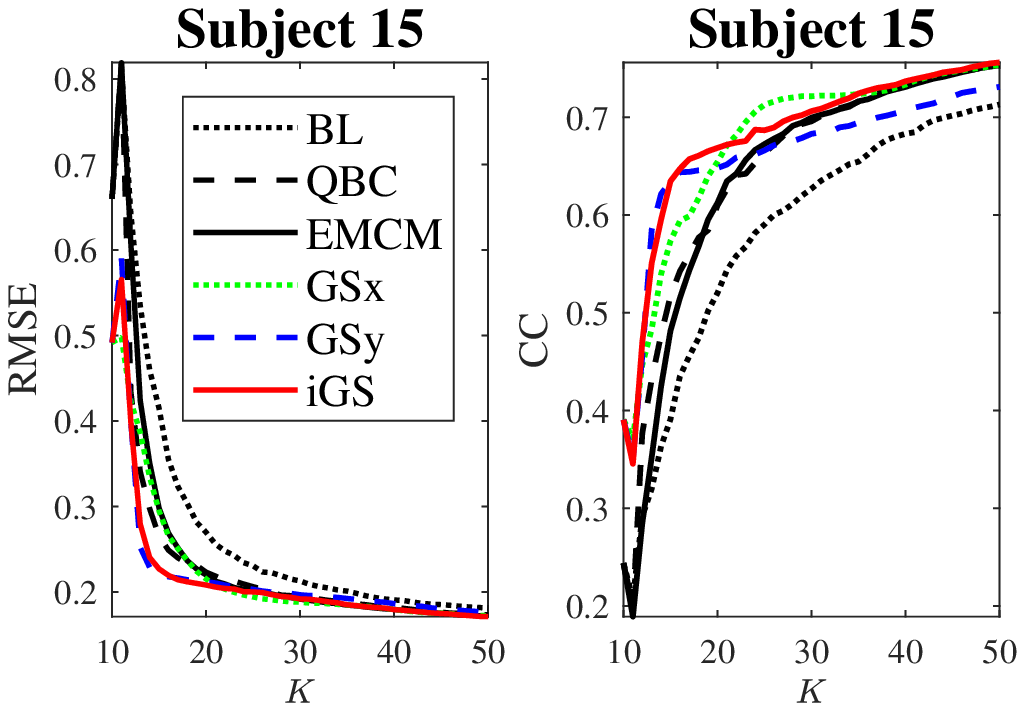}}
\caption{Performances of the six algorithms on the 15 subjects, averaged over 100 runs.} \label{fig:results15}
\end{figure*}

\begin{figure}[!h]\centering
\includegraphics[width=.8\linewidth,clip]{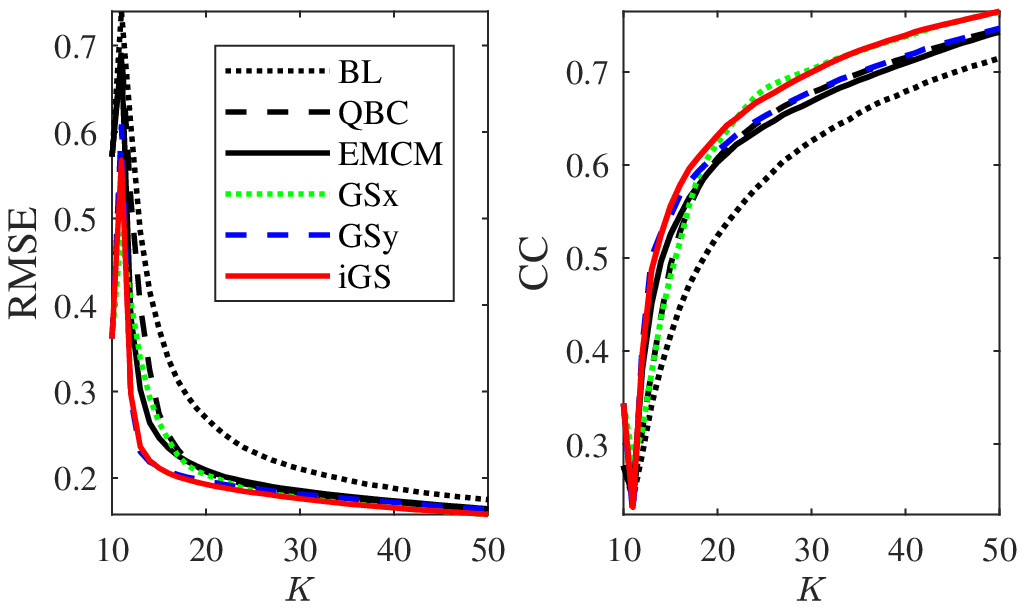}
\caption{Average performances of the six algorithms over the 15 subjects.} \label{fig:avg}
\end{figure}

The RMSEs and CCs for the six algorithms on the 15 subjects, averaged over 100 runs, are shown in Figure~\ref{fig:results15}. The average performances over the 15 subjects are shown in Figure~\ref{fig:avg}. Generally as $K$ increased, all six algorithms achieved better performance (smaller RMSE and larger CC), which is intuitive, because more labeled training samples generally result in a more reliable RR model. \texttt{iGS} achieved the smallest RMSE and largest CC for most subjects.

\begin{figure}[!h]\centering
\subfigure[]{\label{fig:AUC-RMSE}     \includegraphics[width=.96\linewidth,clip]{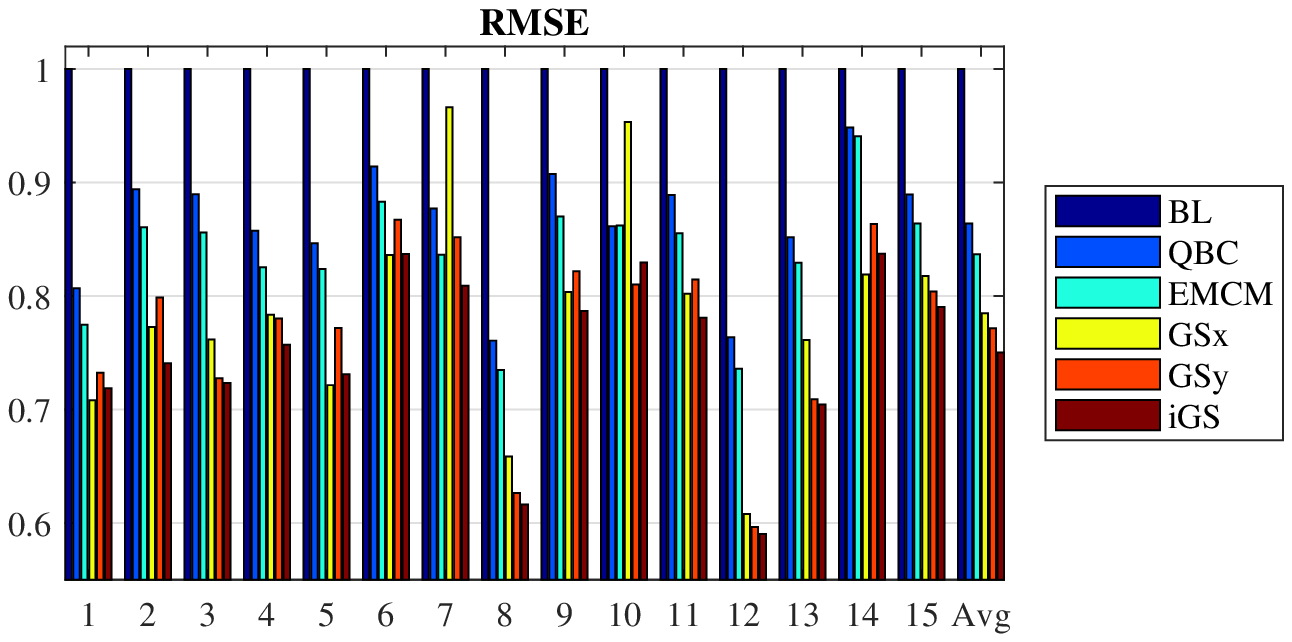}}
\subfigure[]{\label{fig:AUC-CC}     \includegraphics[width=.96\linewidth,clip]{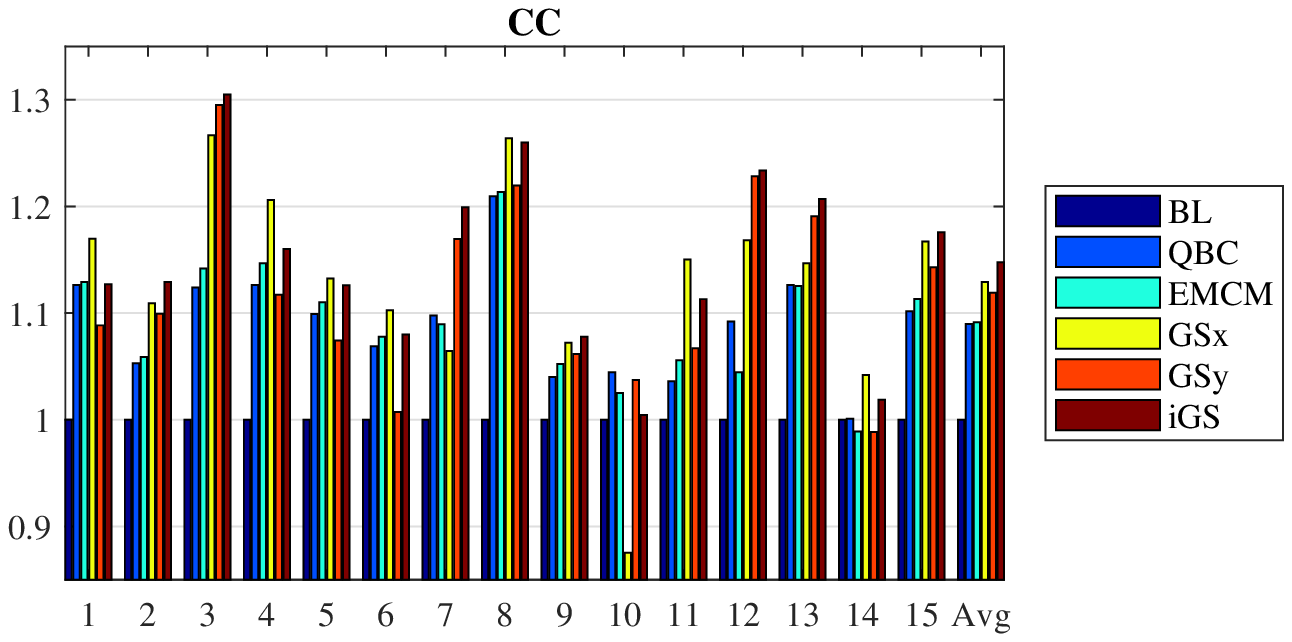}}
\caption{Normalized AUCs of the six algorithms on the 15 subjects. (a) RMSE; (b) CC.} \label{fig:AUC-Driving}
\end{figure}

The AUCs for the RMSEs are shown in Figure~\ref{fig:AUC-RMSE}. Again, for each subject, we used the AUC of \texttt{BL} to normalize the AUCs of the other five algorithms, so the AUC of \texttt{BL} was always 1. As before, a smaller AUC for RMSE indicates a better performance, and a larger AUC for CC indicates a better performance. Figure~\ref{fig:AUC-Driving} shows that:
\begin{enumerate}
\item Each of the five ALR approaches achieved smaller RMSEs than \texttt{BL} on all 15 subjects, and larger CCs than \texttt{BL} on at least 14 subjects, suggesting that they were all effective.
\item On average the performances of the six algorithms, from the best to the worst, was \texttt{iGS} $>$ \texttt{GSy} $\approx$ \texttt{GSx} $>$ \texttt{EMCM} $>$ \texttt{QBC} $>$ \texttt{BL}. \texttt{iGS} combines the advantages of \texttt{GSx} and \texttt{GSy}, and outperformed both of them.
\end{enumerate}
These observations were also confirmed by Table~\ref{tab:ranksDriving}, which shows the detailed ranks of the six approaches on the 15 subjects, according to the AUCs.

\begin{table}[!h] \centering \setlength{\tabcolsep}{1.5mm}
\caption{Ranks of the six approaches on the 15 subjects.}   \label{tab:ranksDriving}
\begin{tabular}{c|c|cccccc}   \hline
   &Subject & \texttt{BL}  & \texttt{QBC} & \texttt{EMCM} & \texttt{GSx} & \texttt{GSy} & \texttt{iGS} \\ \hline
   &1      &6&5&4&1&3&2 \\
   &2      &6&5&4&2&3&1 \\
   &3      &6&5&4&3&2&1 \\
   &4      &6&5&4&3&2&1 \\
   &5      &6&5&4&1&3&2 \\
   &6      &6&5&4&1&3&2 \\
   &7      &6&4&2&5&3&1 \\
RMSE&8     &6&5&4&3&2&1 \\
   &9      &6&5&4&2&3&1 \\
   &10     &6&3&4&5&1&2 \\
   &11     &6&5&4&2&3&1 \\
   &12     &6&5&4&3&2&1  \\
   &13     &6&5&4&3&2&1 \\
   &14     &6&5&4&1&3&2 \\
   &15     &6&5&4&3&2&1  \\
   &\textbf{Average}    &\textbf{6}&\textbf{5}&\textbf{4}&\textbf{3}&\textbf{2}&\textbf{1}\\  \hline
   &1      &6&4&2&1&5&3 \\
   &2      &6&5&4&2&3&1 \\
   &3      &6&5&4&3&2&1 \\
   &4      &6&4&3&1&5&2 \\
   &5      &6&4&3&1&5&2 \\
   &6      &6&4&3&1&5&2 \\
   &7      &6&3&4&5&2&1 \\
CC &8      &6&5&4&1&3&2 \\
   &9      &6&5&4&2&3&1 \\
   &10     &5&1&3&6&2&4 \\
   &11     &6&5&4&1&3&2 \\
   &12     &6&4&5&3&2&1  \\
   &13     &6&4&5&3&2&1 \\
   &14     &4&3&5&1&6&2 \\
   &15     &6&5&4&2&3&1   \\
   &\textbf{Average}   &\textbf{6}&\textbf{5}&\textbf{4}&\textbf{2}&\textbf{3}&\textbf{1}\\ \hline
\end{tabular}
\end{table}
\renewcommand{\baselinestretch}{1.5}

\subsection{Statistical Analysis}

To determine if the differences between different pairs of algorithms were statistically significant, we also performed non-parametric multiple comparison tests on the AUCs using the procedure in Section~\ref{sect:SA}. The $p$-values for the AUCs of RMSEs and CCs are shown in Table~\ref{tab:DunnDriving}, where the statistically significant ones are marked in bold. Table~\ref{tab:DunnDriving} shows that:
\begin{enumerate}
\item All five ALR approaches had statistically significantly better RMSEs and CCs than \texttt{BL}, suggesting that they were effective.
\item Among the three existing ALR approaches, \texttt{GSx} had statistically significantly better RMSEs and CCs than \texttt{QBC} and \texttt{EMCM}.
\item \texttt{GSy} had statistically significantly better RMSE than \texttt{QBC} and \texttt{EMCM}, and statistically significantly better RMSE than \texttt{QBC} and \texttt{GSx}.
\item \texttt{iGS} had statistically significantly better RMSE than all other approaches, and also statistically significantly better CC than all other approaches except \texttt{GSx}.
\end{enumerate}
All these observations were generally consistent with our observations on the 12 UCI and CMU StatLib datasets, demonstrating the effectiveness and robustness of our proposed approaches. Particularly, on average \texttt{iGS} achieved the best performance among the six.

\begin{table}[!h] \centering \setlength{\tabcolsep}{2mm}
\caption{$p$-values of non-parametric multiple comparisons on the AUCs of RMSEs and CCs on EEG-based driver drowsiness estimation.}   \label{tab:DunnDriving}
\begin{tabular}{c|l|ccccc}   \hline
&   & \texttt{BL} &           \texttt{QBC} &     \texttt{EMCM}  & \texttt{GSx} &  \texttt{GSy}   \\ \hline
&\texttt{QBC} & \textbf{.0000} & & & &\\
&\texttt{EMCM} & \textbf{.0000} & \textbf{.0000} & & & \\
RMSE&\texttt{GSx} & \textbf{.0000} & \textbf{.0000} & \textbf{.0000}& \\
&\texttt{GSy} & \textbf{.0000} & \textbf{.0000} & \textbf{.0000} &.0955 & \\
&\texttt{iGS} & \textbf{.0000} & \textbf{.0000} & \textbf{.0000} & \textbf{.0002} &\textbf{.0001} \\   \hline
&\texttt{QBC} & \textbf{.0000} & & & &\\
&\texttt{EMCM} & \textbf{.0000} & .2491 & & & \\
CC&\texttt{GSx} & \textbf{.0000} & \textbf{.0000} & \textbf{.0000}& \\
&\texttt{GSy} & \textbf{.0000} & \textbf{.0080} & .0376 &\textbf{.0065} & \\
&\texttt{iGS} & \textbf{.0000} & \textbf{.0000} & \textbf{.0000} & .0394 &\textbf{.0000} \\   \hline
\end{tabular}
\end{table}

\section{Conclusions and Future Research} \label{sect:conclusions}

Usually a substantial amount of labeled training samples are needed to build an accurate regression model with good generalization ability. However, many times in real-world applications we can collect a large number of unlabeled samples, but labeling them is time-consuming or expensive. ALR is a methodology to select the most beneficial unlabeled samples to label, so that a better regression model can be built from a small number of labeled samples. This paper has proposed two new ALR approaches, inspired by a GS approach in the literature. Extensive experiments on 12 UCI and CMU StatLib datasets, and on 15 subjects on EEG-based driver drowsiness estimation, verified their effectiveness and robustness. Particularly, our proposed \texttt{iGS}, which considers diversity in both input and output spaces, outperformed several existing ALR approaches.

Our future research will extend \texttt{GSx}, \texttt{GSy} and \texttt{iGS} from regression to classification. Additionally, as described in the Introduction, regularization, transfer learning, and active learning can all be used to handle regression problems that do not have enough labeled training data. In this paper we have used regularization and active learning together. In the future we will also study how to integrate transfer learning and active learning for regression problems. Our previous research has integrated transfer learning and active learning for classification problems and achieved promising performance \cite{drwuTNSRE2016,drwuPLOS2013,drwuSMC2017}.

\section*{References} 

\begin{thebibliography}{10}
\expandafter\ifx\csname url\endcsname\relax
  \def\url#1{\texttt{#1}}\fi
\expandafter\ifx\csname urlprefix\endcsname\relax\def\urlprefix{URL }\fi
\expandafter\ifx\csname href\endcsname\relax
  \def\href#1#2{#2} \def\path#1{#1}\fi

\bibitem{drwuICME2010}
D.~Wu, T.~D. Parsons, E.~Mower, S.~S. Narayanan, Speech emotion estimation in
  {3D} space, in: Proc. {IEEE} Int'l Conf. on Multimedia \& Expo ({ICME}),
  Singapore, 2010, pp. 737--742.

\bibitem{drwuInterSpeech2010}
D.~Wu, T.~D. Parsons, S.~S. Narayanan, Acoustic feature analysis in speech
  emotion primitives estimation, in: Proc. {InterSpeech}, Makuhari, Japan,
  2010.

\bibitem{Mehrabian1980}
A.~Mehrabian, Basic Dimensions for a General Psychological Theory: Implications
  for Personality, Social, Environmental, and Developmental Studies,
  Oelgeschlager, Gunn \& Hain, 1980.

\bibitem{Grimm2008}
M.~Grimm, K.~Kroschel, S.~S. Narayanan, The {V}era {A}m {M}ittag {G}erman
  audio-visual emotional speech database, in: Proc. Int'l Conf. on Multimedia
  \& Expo ({ICME}), Hannover, German, 2008, pp. 865--868.

\bibitem{Bradley2007}
M.~M. Bradley, P.~J. Lang, The international affective digitized sounds (2nd
  edition; {IADS}-2): Affective ratings of sounds and instruction manual, Tech.
  Rep. B-3, University of Florida, Gainesville, FL (2007).

\bibitem{drwuTFS2017}
D.~Wu, V.~J. Lawhern, S.~Gordon, B.~J. Lance, C.-T. Lin, Driver drowsiness
  estimation from {EEG} signals using online weighted adaptation regularization
  for regression ({OwARR}), {IEEE} Trans. on Fuzzy Systems 25~(6) (2017)
  1522--1535.

\bibitem{drwuEBMAL2016}
D.~Wu, V.~J. Lawhern, S.~Gordon, B.~J. Lance, C.-T. Lin, Offline {EEG}-based
  driver drowsiness estimation using enhanced batch-mode active learning
  ({EBMAL}) for regression, in: Proc. {IEEE} Int'l Conf. on Systems, Man and
  Cybernetics, Budapest, Hungary, 2016, pp. 730--736.

\bibitem{drwuSMLR2016}
D.~Wu, V.~J. Lawhern, S.~Gordon, B.~J. Lance, C.-T. Lin, Spectral meta-learner
  for regression {(SMLR)} model aggregation: Towards calibrationless
  brain-computer interface ({BCI}), in: Proc. {IEEE} Int'l Conf. on Systems,
  Man and Cybernetics, Budapest, Hungary, 2016, pp. 743--749.

\bibitem{Zou2005}
H.~Zou, T.~Hastie, Regularization and variable selection via the elastic net,
  Journal of the Royal Statistical Society 67~(2) (2005) 301--320.

\bibitem{Hastie2009}
T.~Hastie, R.~Tibshirani, J.~Friedman, The Elements of Statistical Learning,
  Springer, 2009.

\bibitem{Hoerl1970}
A.~E. Hoerl, R.~W. Kennard, Ridge regression: Biased estimation for
  nonorthogonal problems, Technometrics 12~(1) (1970) 55--67.

\bibitem{Tibshirani1996}
R.~Tibshirani, Regression shrinkage and selection via the lasso, Journal of the
  Royal Statistical Society 58~(1) (1996) 267--288.

\bibitem{Pan2010}
S.~J. Pan, Q.~Yang, A survey on transfer learning, {IEEE} Trans. on Knowledge
  and Data Engineering 22~(10) (2010) 1345--1359.

\bibitem{drwuaBCI2015}
D.~Wu, C.-H. Chuang, C.-T. Lin, Online driver's drowsiness estimation using
  domain adaptation with model fusion, in: Proc. Int'l Conf. on Affective
  Computing and Intelligent Interaction, Xi'an, China, 2015, pp. 904--910.

\bibitem{Settles2009}
B.~Settles, Active learning literature survey, Computer Sciences Technical
  Report 1648, University of Wisconsin--Madison (2009).

\bibitem{Sugiyama2009}
M.~Sugiyama, S.~Nakajima, Pool-based active learning in approximate linear
  regression, Machine Learning 75~(3) (2009) 249--274.

\bibitem{Burbidge2007}
R.~Burbidge, J.~J. Rowland, R.~D. King, Active learning for regression based on
  query by committee, Lecture Notes in Computer Science 4881 (2007) 209--218.

\bibitem{Cai2013}
W.~Cai, Y.~Zhang, J.~Zhou, Maximizing expected model change for active learning
  in regression, in: Proc. {IEEE} 13th Int'l. Conf. on Data Mining, Dallas, TX,
  2013.

\bibitem{Yu2010}
H.~Yu, S.~Kim, Passive sampling for regression, in: {IEEE} Int'l. Conf. on Data
  Mining, Sydney, Australia, 2010, pp. 1151--1156.

\bibitem{Cai2017}
W.~Cai, M.~Zhang, Y.~Zhang, Batch mode active learning for regression with
  expected model change, {IEEE} Trans. on Neural Networks and Learning Systems
  28~(7) (2017) 1668--1681.

\bibitem{drwuSAL2018}
D.~Wu, Pool-based sequential active learning for regression, {IEEE} Trans. on
  Neural Networks and Learning Systems, 2018, submitted.

\bibitem{RayChaudhuri1995}
T.~RayChaudhuri, L.~Hamey, Minimisation of data collection by active learning,
  in: Proc. {IEEE} Int'l. Conf. on Neural Networks, Vol.~3, Perth, Australia,
  1995, pp. 1338--1341.

\bibitem{Dunn1961}
O.~Dunn, Multiple comparisons among means, Journal of the American Statistical
  Association 56 (1961) 62--64.

\bibitem{Dunn1964}
O.~Dunn, Multiple comparisons using rank sums, Technometrics 6 (1964) 214--252.

\bibitem{Benjamini1995}
Y.~Benjamini, Y.~Hochberg, Controlling the false discovery rate: A practical
  and powerful approach to multiple testing, Journal of the Royal Statistical
  Society, Series B (Methodological) 57 (1995) 289--300.

\bibitem{Chuang2012}
S.-W. Chuang, L.-W. Ko, Y.-P. Lin, R.-S. Huang, T.-P. Jung, C.-T. Lin,
  Co-modulatory spectral changes in independent brain processes are correlated
  with task performance, Neuroimage 62 (2012) 1469--1477.

\bibitem{Chuang2014}
C.-H. Chuang, L.-W. Ko, T.-P. Jung, C.-T. Lin, Kinesthesia in a
  sustained-attention driving task, Neuroimage 91 (2014) 187--202.

\bibitem{Delorme2004}
A.~Delorme, S.~Makeig, {EEGLAB}: an open source toolbox for analysis of
  single-trial {EEG} dynamics including independent component analysis, Journal
  of Neuroscience Methods 134 (2004) 9--21.

\bibitem{Welch1967}
P.~Welch, The use of fast {F}ourier transform for the estimation of power
  spectra: A method based on time averaging over short, modified periodograms,
  {IEEE} Trans. on Audio Electroacoustics 15 (1967) 70--73.

\bibitem{drwuTNSRE2016}
D.~Wu, V.~J. Lawhern, W.~D. Hairston, B.~J. Lance, Switching {EEG} headsets
  made easy: {Reducing} offline calibration effort using active weighted
  adaptation regularization, {IEEE} Trans. on Neural Systems and Rehabilitation
  Engineering 24~(11) (2016) 1125--1137.

\bibitem{drwuPLOS2013}
D.~Wu, B.~J. Lance, T.~D. Parsons, Collaborative filtering for brain-computer
  interaction using transfer learning and active class selection, {PLoS ONE}.

\bibitem{drwuSMC2017}
D.~Wu, Active semi-supervised transfer learning ({ASTL}) for offline {BCI}
  calibration, in: Proc. {IEEE} Int'l. Conf. on Systems, Man and Cybernetics,
  Banff, Canada, 2017.

\end{thebibliography}

\end{document}